\documentclass[acmlarge]{acmart}
\usepackage{pifont}

\usepackage{amssymb}
\usepackage{amsmath}
\usepackage[ruled,linesnumbered]{algorithm2e} %
\usepackage{hyperref}
\usepackage{diagbox}
\usepackage{amsthm}
\usepackage{enumitem}
\usepackage{multirow}
\usepackage{booktabs}
\usepackage{subfig}
\usepackage{amsmath,amsfonts}
\usepackage{algorithmic}
\usepackage{array}
\usepackage{subfig}
\usepackage{multirow}
\usepackage{pdflscape}
[caption=false,font=normalsize,labelfont=sf,textfont=sf]
\usepackage{booktabs} 
\usepackage{enumerate}
\usepackage{bm}
\usepackage{lipsum}
\usepackage{verbatim}
\usepackage{adjustbox}
\usepackage[graphicx]{realboxes}
\usepackage{rotating}
\usepackage{textcomp}
\usepackage{stfloats}
\usepackage{url}
\usepackage{verbatim}
\usepackage{graphicx}
\usepackage{rotating}
\usepackage{balance}

\newcommand{\mypara}[1]{{\noindent\textbf{#1}}}
\newcommand{\DiffBot}{\textsc{DiffBot}\xspace}
\newcommand{\RoBCtrl}{\textsc{RoBCtrl}\xspace}
\newcommand{\RoBCtrlSA}{\textsc{RoBCtrl+SA}\xspace}
\theoremstyle{definition}
\newtheorem{definition}{Definition}[section]

\AtBeginDocument{%
  }

\setcopyright{acmlicensed}
\copyrightyear{2018}
\acmYear{2018}
\acmDOI{XXXXXXX.XXXXXXX}
\acmConference[Conference acronym 'XX]{Make sure to enter the correct
  conference title from your rights confirmation email}{June 03--05,
  2018}{Woodstock, NY}
\acmISBN{978-1-4503-XXXX-X/2018/06}

\begin{document}
\title{\RoBCtrl: Attacking GNN-Based Social Bot  
 Detectors via Reinforced Manipulation of Bots Control Interaction}

\author{Yingguang Yang}
\email{dao@mail.ustc.edu.cn}
\affiliation{%
  \institution{University of Science and Technology of China}
  \city{Hefei}
  \state{Anhui}
  \country{China}
}
\author{Xianghua Zeng}
\email{zengxianghua@buaa.edu.cn}
\affiliation{%
  \institution{Beihang University}
  \city{Beijing}
  \country{China}
}
\author{Qi Wu}
\email{qiwu4512@mail.ustc.edu.cn}
\affiliation{%
  \institution{University of Science and Technology of China}
  \city{Hefei}
  \state{Anhui}
  \country{China}
}

\author{Hao Peng}
\authornote{This is the corresponding author.}
\email{penghao@buaa.edu.cn}
\affiliation{%
  \institution{Beihang University}
  \city{Beijing}
  \country{China}
}

\author{Yutong Xia}
\email{yutong.xia@u.nus.edu}
\affiliation{%
 \institution{National University of Singapore}
 \city{Singapore}
 \country{Singapore}}

\author{Hao Liu}
\email{rcdchao@mail.ustc.edu.cn}
\affiliation{%
  \institution{University of Science and Technology of China}
  \city{Hefei}
  \state{Anhui}
  \country{China}
}

\author{Bin Chong}
\authornotemark[1]
\affiliation{%
  \institution{Peking University}
  \city{Beijing}
  \country{China}}
\email{chongbin@pku.edu.cn}

\author{Philip S. Yu}
\email{psyu@uic.edu}
\affiliation{%
  \institution{University of Illinois Chicago}
  \city{Chicago}
  \country{USA}}


\begin{abstract}
Social networks have become a crucial source of real-time information for individuals. 
The influence of social bots within these platforms has garnered considerable attention from researchers, leading to the development of numerous detection technologies.
However, the vulnerability and robustness of these detection methods is still underexplored. 
Existing Graph Neural Network (GNN)-based methods cannot be directly applied due to the issues of \textit{limited control over social agents}, \textit{the black-box nature of bot detectors}, and \textit{the heterogeneity of bots}. 
To address these challenges, this paper proposes the first adversarial multi-agent \underline{R}einf\underline{o}rcement learning framework for social \underline{B}ot \underline{c}on\underline{tr}o\underline{l} attacks (\textbf{\RoBCtrl}) targeting GNN-based social bot detectors.
Specifically, we use a diffusion model to generate high-fidelity bot accounts by reconstructing existing account data with minor modifications, thereby evading detection on social platforms. 
To the best of our knowledge, this is the first application of diffusion models to mimic the behavior of evolving social bots effectively.  
We then employ a Multi-Agent Reinforcement Learning (MARL) method to simulate bots' adversarial behavior. 
We categorize social accounts based on their influence and budget.
Different agents are then employed to control bot accounts across various categories, optimizing the attachment strategy through reinforcement learning. 
Additionally, a hierarchical state abstraction based on structural entropy is designed to accelerate the reinforcement learning. 
Extensive experiments on social bot detection datasets demonstrate that our framework can effectively undermine the performance of GNN-based detectors. 
\end{abstract}



\received{February 2025}

\maketitle

\section{INTRODUCTION}
Social media platforms have become indispensable for communication and information dissemination, fostering unprecedented connectivity. 
This ubiquity has also contributed to the growing presence of social bots over the past decade \cite{cresci2020decade}. 
These social bots, designed to mimic human behavior, are often deployed for malicious purposes, including spreading misinformation \cite{wang2020fake,lu2020gcan}, spreading propaganda \cite{boshmaf2011socialbot,chaudhari2021propaganda}, and election interference \cite{keller2019social,ferrara2020characterizing}. 
To mitigate the detrimental impacts of social bots on economic and political development, a substantial body of work has been devoted to effectively detecting bots within social networks. 
While early approaches relied on feature engineering (e.g., features and metadata extraction \cite{cresci2016dna,d2015real,yang2020scalable}), the advent of Deep Learning (DL) has significantly enriched this field. Recent bot detectors leverage deep neural networks, including Recurrent Neural Networks~\cite{kudugunta2018deep,wei2019twitter}, self-supervised learning~\cite{feng2021satar,yang2023fedack}, and, most notably, Graph Neural Networks (GNNs) ~\cite{breuer2020friend,dou2020enhancing,feng2022heterogeneity,feng2021botrgcn,liu2018heterogeneous,yang2022rosgas}. 
Among these, GNN-based methods have emerged as the most advanced approach for identifying social bots at scale, offering exceptional capabilities for modeling complex and dynamic interactions within social networks. 
These advances are pivotal combating against malicious bots, enabling more robust and scalable detection strategies.

Although promising, the vulnerability and robustness of the abovementioned detection approaches remain largely underexplored. In practical scenarios, most existing works \cite{mendoza2020bots,yang2020scalable,moghaddam2022friendship,li2022botfinder,shi2023rf,wu2023heterophily,peng2024unsupervised,he2024dynamicity,yang2024sebot} often overlook the fact that malicious bots actively interact with regular users to disguise themselves, thus reducing the likelihood of detection.
This oversight limits the effectiveness of these methods in combating social bots with \textit{evolutionary} traits \cite{cresci2020decade,yang2013empirical}. 
Few studies have focused on investigating adversarial behaviors or modeling the diversity of different types of bots. 
As a result, emerging bots can easily evade from detection techniques, placing social platforms in a passive position within constantly evolving social environments. 
Therefore, it is imperative to study the adversarial attack patterns of social bots and further enhance the robustness of detection models.

Although recent studies~\cite{sun2022adversarial} have provided valuable insights regarding the robustness of GNNs and introduced plausible methods to simulate the adversarial actions of social bots, these approaches cannot be directly applied to bot detection because of the following limitations. 
i) \textbf{\textit{Limited control of social agents}}. 
Previous adversarial attack strategies on GNNs assume unrestricted disturbance across all nodes and edges. 
However, this assumption is unrealistic because bot developers only have limited control over social accounts and cannot access the entire network structure. 
Effective simulation requires models that operate under realistic constraints that accurately reflect actual agent control in social systems.
ii) \textbf{\textit{Black-box nature of bot detectors}}. Methods that target GNN-based bot detectors must treat the bot detectors as black boxes.
The lack of knowledge about the model architecture renders the prevailing gradient-based attack methodologies \cite{chen2018fast} impractical in real-world settings. 
iii) \textbf{\textit{Bot heterogeneity}}. 
These approaches do not account for the heterogeneity inherent in social bots, including variations in bot types, optimization functions, adversarial budgets, and varying risk preferences. 
Ignoring these factors reduces the effectiveness of adversarial strategies, as the nuanced interactions between different bot types and their environments are not adequately modeled.
These limitations highlight the need for more tailored frameworks that account for the realistic dynamics of adversarial agents, the opaque nature of GNN-based detectors, and the heterogeneity of social bots to ensure robust and practical advancements in this domain.


This paper presents \underline{R}einf\underline{o}rcement learning social \underline{B}ot \underline{C}on\underline{tr}o\underline{l} framework (\textbf{\RoBCtrl}), an adversarial attack framework designed to address the key limitations mentioned above in existing GNN-based social bot detection methods.
Our framework integrates a diffusion model for feature generation and reinforcement learning-based mechanism to control bot interactions. The key insight is that diffusion models naturally align with and can effectively simulate the behavior of even the most advanced, evolving social bots.
The process of gradually removing Gaussian noise and then incrementally adding noise to generate approximate samples closely mirrors how evolving bots steal and rebuild user information with subtle modifications. 
We then employ a multi-agent reinforcement learning (MARL) method, leveraging its exceptional capability to operate in black-box scenarios---without knowledge of the detector’s model structure---to replicate replicate the adversarial behavior of social bots while incorporating factors such as individual social influence and adversarial budget. 
To capture the heterogeneity of bots, we model different types of bots based on their characteristics and assign distinct agents to represent and manipulate each type of bots.
Reinforcement learning is then employed to optimize the attack strategy. 
To further enhance computational efficiency, we incorporate hierarchical state abstraction into the framework. 
Specifically, by constructing an optimal encoding tree for environment states in MARL based on structural information principles\cite{li2016structural},  this approach groups similar environment states together to reduce complexity. 
Extensive experiments are conducted on social bot detection datasets and show that our method can effectively undermine the ecosystems of social networks and degenerate GNN-based detectors' performance. 
Additional experiments on two datasets demonstrate that state abstraction can reduce the time required for a successful attacks.
Our findings reveal the drawbacks of the existing detection methods and highlight the demand for more robust and reliable detection methods to uphold the content security of web-based applications (e.g.Twitter) within a safe and healthy ecosystem.

Our contributions can be summarized as follows:
\begin{itemize}[leftmargin=*]
    \item \textbf{Diffusive Feature Simulation.} 
    To the best of our knowledge, we propose the first diffusion model for feature generation that simulates how bot account features are created, effectively mimicking the behavior of \textit{evolving} social bots.
    \item \textbf{Collaborative Bot Control.} 
    We develop a MARL-based social bot control framework to imitate the collaborative attacking behavior by strategically manipulating the interactions of different bot types.
    \item \textbf{New Vulnerability Insights.} 
    We provide a new perspective on the vulnerability of social bot detection methods and open new avenues for further research in this area.
    \item \textbf{Extensive Empirical Evidences.} Extensive experiments on social bot detection datasets demonstrate that our method effectively disrupts social network ecosystems and degrades GNN-based detectors. 
\end{itemize}

The structure of this paper is as follows. 
Section \ref{sec:relatedwork} provides an overview of related works.
Section \ref{sec:Preliminaries} outlines the background and fundamental concepts of our work. 
we describe the technical details of the proposed framework, named \RoBCtrl in Section \ref{sec:Methodology}. 
Section \ref{sec:Evaluation} presents the experimental setup and discusses the experiment’s results. 
Finally, we conclude the paper in Section \ref{sec:Conclusion}.
\section{RELATED WORK}
\label{sec:relatedwork}

\subsection{GNN-based Social Bot Detection} 
A social network usually consists of rich information such as social familiarity \cite{dey2018assessing,dey2019assessing}, attribution similarity \cite{peng2018anomalous}, and user interaction \cite{viswanath2009evolution}. 
The construction of graphs based on social networks is well suited for detecting social bots by applying recently popularized GNN methods. 
GNNs decelerate the imitation and evolutionary velocity of social bots. 
The initial effort \cite{ali2019detect} to employ GNN for social bot detection merely integrates the GNN with multi-layered perception and belief propagation.
In the subsequent works, heterogeneous graphs are constructed to model the influence between users. 
The pre-training language model is utilized to extract the original node features. 
Such features are then used to learn node representations after aggregating by the personalized GCN model \cite{yang2022rosgas}, R-GCN model \cite{feng2021botrgcn}, or relational graph transformer \cite{feng2022heterogeneity}. 
Some studies explore the combination of graph-based and text-based methods \cite{guo2021social,lei2022bic} or devise new GNN architectures by considering network heterogeneity \cite{feng2022heterogeneity,yang2023fedack}. 
Notably, there has been a recent shift from individual-level detection to community-level detection \cite{tan2023botpercent,liu2023botmoe}, owing to the release of Twibot-22 \cite{feng2022twibot}, a large-scale social bot dataset.  
Model robustness is not the primary focus of these works; thus, they can hardly tackle ever-growing bot evolution with upgraded types of camouflages and attacks.

\subsection{Adversarial Attacks on GNNs}
While successful in graph mining tasks, GNNs are shown to be vulnerable to adversarial attacks.
Attack methods can alter the graph structure and node attributes, leading to non-negligible performance degradation of GNNs \cite{zugner2018adversarial,wu2019adversarial}. 
Attacks can be classified into two categories: 
(i) \textit{node-specific attacks}, e.g., target evasion attack \cite{wu2019adversarial,zugner2018adversarial}, aiming to deceive GNNs to misclassify specific target nodes; 
(ii) \textit{ nontarget attacks } \cite{dai2018adversarial,sun2020adversarial}, are designed to decrease the accuracy of GNNs across a graph. 
They can be implemented by heuristic approaches \cite{zou2021tdgia}, or reinforcement learning for search optimal attack solutions \cite{dai2018adversarial,sun2020adversarial}. 
These methods perturb the graph by adding/removing edge structures \cite{dai2018adversarial, daniel2019adversarial}, modifying node features \cite{zugner2018adversarial, finkelshtein2022single} or injecting nodes under the given budget. 
However, none is exclusively designated for social bot detectors, without particular consideration of bot characteristics such as influence, adversarial budget, etc.  
More importantly, the graph structure and model architecture are invisible to the GNN-based social bot detectors, making the conventional attacks (e.g., gradient-based attacks) in generic GNNs useless in the scenario of bot detection.

\subsection{Adversarial Social Bot Detection}
The adversarial bots have not yet been well studied. 
The first adversarial detection \cite{cresci2019better} focused on using static snapshots of their characteristics. 
It adopts an evolution optimization algorithm to find adversarial permutations from a fixed social bot activity sequence and examines if they can improve bot detection accuracy. However, these permutations only partially understand the behavior of social bots and thus cannot capture the full temporal dynamicity. 
A follow-up work \cite{le2022socialbots} formulates the social bot behavior as a Markov Decision Process (MDP) and proposes a reinforcement learning framework to optimize the adversarial goals on real-life networks.
While these works pioneered adversarial learning into the task of social bot detection, the applied detection models are relatively primitive and thus have massive potential for improvement. 
This paper makes the first attempt to exploit the robustness of GNN-based social bot detectors through a diffusive generation model and MARL-based framework. 
We endeavor to concurrently target a cohort of users within a unified offensive framework to undermine the overall efficacy of detection mechanisms. 
This objective diverges from the premises and operational scopes of the two most recent works \cite{wang2023attacking,wang2023my}, which aim their offensive measures at isolated users or nodes to alter the detection outcomes specific to those singular entities. 
Moreover, our approach additionally models the evolutionary progression of controlled social bot features, providing a new dimension in social bot adversarial attacks.
\section{BACKGROUND AND PRELIMINARIES}
\label{sec:Preliminaries}
In this section, we first summarize the problems and challenges of social bot detection. 
Then, we define GNN-based social bot detection and introduce the adversarial attack objective. 
The comprehensive list of the primary symbols used throughout this paper is presented in Table \ref{tab: notation}.

\begin{table*}[t]
\aboverulesep=0ex
\belowrulesep=0ex
\caption{List of symbols and definitions.} 
\centering
\begin{tabular}{r|l} 
\toprule
\hline
\textbf{Symbol} & \textbf{Definition} \\
\hline
$\mathcal{G}$ & Homogeneous Graph.\\

$V;E;e_{ij}$ & Vertex set; Edge set of Homogeneous Graph; Edge between $v_i$ and $v_j$.      \\

$X;Y;v_i;y_i$ & Feature set; Label set; The $i$-th user in graph $\mathcal{G}$; The $i$-th label of user $v_i$.\\
$h_v^{(l)};v'$ & The representation vector of $v$ after $l$-th convolution layer; neighbor of vertex $v$.\\

$\sigma;AGG;\mathcal{L}$ & Activation function; Aggregate function; Loss function.  \\
$f_\theta;\mathbb{I}$& The model parameters of the bot classifier; Identification function.\\
$V_T;V_g;E_c$ & Target user set; Generate user set; Generated edge set.\\

$\Delta_v;\Delta_e$ & Budget for the number of nodes and edges generated during the attack. \\
$\mathcal{G}'$ & Contaminated graph data after attack. \\
\hline
$\mathcal{N};\beta_t; t $& Gaussian distribution; Gaussian  Noise scale; $t$-thTime step.\\
$T;\mu_\theta,\Sigma_\theta$ & The final time step; Parameters of Forward Propagation Neural Network.\\
$\boldsymbol{I};\mathbb{E};\mathcal{U}$ & Identity matrix; Mathematical Expectations; Uniform distribution.\\
$\hat{x}_{t}$ & A sample generated at time step t.\\

\hline
$\mathbb{N};N$& Reinforcement learning agents set; Total number of agents.\\
$\mathcal{S};\mathcal{A}^i$ & The state space; The action space of the $i$-th agent.\\
$ s;a $ & A specific state and action.\\
$\mathcal{P};R$ &The state transactions; The reward function.\\

$ \gamma; e()$ & The discount factor of long term reward accumulation; Embedding function.\\
$ U^i_c $ & The set of users controlled by the $i$-th agent.\\
$ a^i_t(u,v)  $ & At time step t, agent i adds edges between $u$ and $v$.\\
$ Q;W$ & Accumulated reward value; Parameters of a fully connected network.\\

$h_{u,t}  $ & The representation vector obtained for node $u$ at time step $t$.\\
$h_{\mathcal{G}_t}  $ & The obtained representation vector for attacked graph $\mathcal {G}_t $ at time step $t $,\\
\hline
$\mathcal{T}; \lambda$ & Encoding tree; The root node of the encoding tree. \\
$ v_\tau;T_\alpha$ & Non-root node on encoding tree; Label of node $\alpha$. \\
$\mathcal{V}_\tau  $ & A subset node from $V$ and start from $v$ as subtree root node in the encoding tree.\\
$v_\tau^+  $ & The immediate predecessor of $v_\tau$.\\
$vol(v_\tau);vol(v_\tau^+)$ & the sum of the degrees of nodes in $v_\tau,v_\tau^+$.\\
$g_{v_\tau}  $ &The number of edges connecting nodes inside $\mathcal{V}_\tau$ with those outside.\\
$\mathcal{H}^\mathcal{T}  $ & Structural Entropy value when the encoding tree is $\mathcal{T} $.\\
$\mathcal{T}_U^*$ & Optimal encoding tree partition.\\
$\mathcal{H}^{\mathcal{T}^*_U}(\mathcal{G}_t;\lambda_i \rightarrow \nu_{i,j})$  & The optimal path entropy from node $\ lambda_i $to $\ nu_ {i, j} $ under the optimal $\mathcal{T}_U^*$.\\
$P;P_{i,j}  $ & Pooling matrix and the $i$-th row and $j$-th column values in the pooling matrix.\\
\hline
\bottomrule
\end{tabular}
\label{tab: notation}
\end{table*} 

\subsection{Problem and Challenges}
To study the robustness of social bot detectors based on GNNs and successfully attack them, three main research challenges must be solved:

\noindent
\textbf{Challenge 1: Modeling different types of social bots} 

Existing approaches often treat social bots as homogeneous entities, ignoring their inherent heterogeneity in objectives, adversarial capabilities, and behavioral patterns. 
For instance, bots may vary in their evasion strategies (e.g., mimicking human interaction patterns vs. stealing user profiles) or adversarial budgets (e.g., limits to their control over network connections). 
Current taxonomies fail to capture these nuances, relying on simplistic categorizations based on static features or predefined rules. 
This limitation hinders the development of robust detection systems as evolving bots dynamically adapt their tactics to evade detection.
A key challenge lies in designing a flexible framework that models various types of bots by incorporating dynamic behavioral traits, risk preferences, and adversarial goals, thereby enabling the simulation of heterogeneous bot populations in adversarial environments.

\noindent
\textbf{Challenge 2: Implementing attacks on GNN-based social bot detectors under black-box conditions} 

Prevailing strategies for adversarial attacks on GNNs, such as gradient-based perturbation methods, assume full access to the target model’s architecture and parameters (white-box setting). 
However, in real-world scenarios, bot operators have no knowledge of the internal workings of detection systems, which is a black-box setting. 
This renders gradient-guided attacks impractical because the adversaries cannot compute loss gradients or directly manipulate model outputs. 
Furthermore, existing black-box attack methods for graphs often rely on excessive queries to the target model, which risks detection and suspension by platform defenses. 
Hence, a key challenge is to devise query-efficient attack strategies that can infer the detector’s decision boundaries without access to explicit gradient information while also adhering to realistic constraints on observable feedback (e.g., suspension alerts) and avoiding detectable query patterns.

\noindent
\textbf{Challenge 3: Successful attacks using a limited control budget and optimized attack process} 

Bot developers operate under strict resource constraints because they control only a small subset of nodes (fake accounts) and edges (social interactions) within a vast social network. 
Traditional adversarial attack methods assume an unrestricted ability to modify graph structures or node features and are hence infeasible in this context. 
Optimizing adversarial actions—such as rewriting stolen profile content or strategically connecting to human users—requires balancing immediate evasion rewards against long-term risks (e.g., over-perturbation, which raises suspicion).
This challenge is further compounded by the need to coordinate heterogeneous bots with varying budgets and objectives. 
Effective solutions must integrate constrained optimization techniques with multi-agent collaboration mechanisms while also reducing computational complexity in large-scale social networks.

\subsection{GNN-based Social Bot Detection}
We formulate the problem of attacking social bot detectors one of attacking a GNN on a user account-homogeneous graph. The user graph is defined as $\mathcal{G}=\{V, E, X, Y\}$, where $V=\{v_1, \dots, v_i\}$ is user set, and the edge $e_{ij}=(v_i,v_j)\in E$ indicates that the user $v_i$ has a relationship/interaction with the user $v_j$. 
In addition, $X$ represents the feature matrices of the user nodes.
According to \cite{yang2023fedack,yang2022rosgas}, user features are composed of metadata features or text representation. Moreover, $y_i \in Y$ represents the label of $v_i \in V$, where 1 represents a social bot, and 0 represents a benign user.
After the user graph has been constructed using the relationships and interactions described in the social platform, a general GNN framework based on $\mathcal{G}$ can be applied to detect social bots. 
A GNN aggregates its neighbors' information recursively to learn the representation for social media account $v$ as follows:
\begin{equation}
    h^{(l)}_v=\sigma(h^{(l-1)}_v\oplus AGG^{(l)}(h^{(l-1)}_{v'}, (v, v')\in E)),
\end{equation}
where $l$ denotes the number of layers of the GNN. 
The GNN aggregator, denoted by $AGG$, is used to aggregate the embeddings of the neighbors. 
This can be done using attention, meaning, or summation. Operator $\oplus$ is the operator that combines the $v$ and its neighbours' information, e.g., using concatenation or summation.
To classify a user $v\in V$ as benign or bot, a GNN classifier $f_\theta$ takes the $\mathcal{G}$ as input. 
It maps $v\in V$ to $y\in (0,1)$ after feeding the $h_v$ from the last MLP layer to the softmax layer. 
Then, the GNN classifier is trained on partially labeled nodes with the following cross-entropy loss:
\begin{equation}
    \mathcal{L}(\mathcal{G},f_\theta) = \sum_{v_i\in V} -log(y_i\cdot\sigma(f_\theta(E,X)_i))
\end{equation}

\subsection{Type-differentiating Adversarial Attacks}
\label{sec:3.2}
Multiple studies \cite{cresci2020decade,yang2013empirical} have demonstrated the existence of distinct populations and generations of social bots on social platforms. 
Specifically, a population or generation of social bots is typically deployed using the same creation and control technologies, and these bots often exhibit signs of coordinated behavior to rapidly achieve a common goal.
Because of the dissimilarities among social bots,
the efforts required for them to achieve malicious objectives while evading detection vary, resulting in differing demands on their budget for attacking a detector and reducing its performance.
Hence, to discern their unique operational dynamics, various social bots must be simulated.
In practice, we simulate three different types of social bot accounts:

\mypara{Automated Bots} have minimal personal information and few social connections, and they primarily make automated posts containing written content~\cite{yang2013empirical}. 

\mypara{Cyborg Bots} possess more detailed personal information and actively establish social connections with other users ~\cite{cresci2020decade}.

\mypara{Evolving Bots} steal information from real users and exhibit behavior that closely resembles that of human-controlled accounts \cite{ferrara2016rise,cresci2017paradigm}. They often publish a large volume of neutral messages interspersed with a small amount of malicious content, sometimes generated by generative models to better disguise themselves.

\mypara{Scenario.} There are two typical attack scenarios: poisoning/training time attack and evasion/testing time attack. In the poisoning attacks, the classifier $f$ employs the contaminated graph for training. 
By contrast, the adversarial examples used in evasion attacks are incorporated into the testing samples after $f$ has been trained on a clean graph. 
In the context of this paper, we assume that the target GNN is pre-trained on clean data, i.e., the training data have not been poisoned by the adversary. 

\mypara{Attack hypothesis.}  
It is reasonable to assume that the attacker has access to all post data, user metadata, and friendship information within the social network, as well as the labels of some users. However, the attacker can \textbf{only modify} the interaction data (edges) between the controlled bots and other users. 
All attacks are assumed to operate in a \textit{black-box} setting, without knowledge of the detector's architecture and model parameters. 
Hence, the attack task can be regarded as a black-box evasive structural attack on the GNN-based node classification. 


\mypara{Objective.} 
The objective of the attack is to cooperatively manipulate controlled types of bots so that they interact with other users, thereby affecting the classification results of a group of target accounts. 
We formally define our attack objective as follows:
\begin{equation}
\begin{split}
    &\mathop{\max}\limits_{\mathcal{G'}} \sum_{ v \in V_T} \mathbb{I}(f_{\theta^*}(\mathcal{G}')_v\neq y_v) \\
    s.t.\ {\theta^*} = &\mathop{\arg\min}\limits_{\theta} \mathcal{L}(\mathcal{G},f_\theta),\  
    |V_g| \leq \Delta_v, |E_c| \leq \Delta_e,
\end{split}
\end{equation}
where $V_g$, $E_c$, and $V_T$ denote the sets of generated users, manipulated edges, and target accounts, respectively. In addition, $\mathcal{G}$ represents the clean graph, while $\mathcal{G}'$ represents the perturbed graph, where $\mathcal{G}' = \mathcal{G} \cup V_g \cup E_c $. 
Here, $\Delta_v$ and $\Delta_e$ represent the budgets of controlled users and modified edges. 
The adversarial objective is to maximize the degree of misclassification of the targeted accounts.


\section{METHODOLOGY}
\label{sec:Methodology} 
To effectively attack adversarial social bot detectors, we propose a bot generation model called \underline{Diff}usion \underline{Bot} (\DiffBot) and a bot control framework called \RoBCtrl, detailed in \S~\ref{sec:diffbot} and \S~\ref{sec:robctrl}, respectively. 
Specifically, recognizing that social media platforms host various types and generations of social bots, we first simulate automated and cyborg bots by identifying and utilizing controlled accounts from existing data while mimicking complex evolving bots using a diffusion model. 
Second, we introduce a MARL-based method to attack graph-based detectors by manipulating graph structures.
Additionally, we introduce a state abstraction method to accelerate the reinforcement learning-based attack process.
Details of the generation of evolving bots in \DiffBot and the MARL-based detector attack in \RoBCtrl are presented below.

\subsection{Diffusion Bot Generation Model}\label{sec:diffbot}
We present a novel \underline{Diff}usion \underline{Bot} generation model (\textbf{\DiffBot}) to generate fake data using existing account data from normal users. 
The purpose of \DiffBot is to simulate how attackers construct evolving bot accounts on social platforms. These bots steal and imitate the information and behavior of normal users. 
To enhance the simulation's authenticity, we take the initial features of real accounts as the starting point. 
We then continuously add noise in the forward process and iteratively learn to reconstruct the user's initial features through denoising training.
By modeling complex user initial features via this iterative denoising training, we aim to effectively mimic real user data.
\subsubsection{Forward and Reverse Processes} 
\DiffBot has two critical processes: 
1) a forward process that corrupts users' initial features by adding Gaussian noises step by step, and 2) a reverse process that gradually learns to denoise and output the generated features.

\mypara{Forward Process.} Given a user with the initial feature $x_u$ consisting of the user's metadata and text data, we set $x_0=x_u$ as the initial state and parameterize the transition by
\begin{equation}
    q(x_t|x_{t-1})= \mathcal{N}(x_t;\sqrt{1-\beta_t}x_{t-1},\beta_t\boldsymbol{I}),
\end{equation}
where $\beta_t\in (0,1)$ controls the Gaussian noise scales added at each step $t$. Using the reparameterization trick \cite{ho2020denoising} and the additivity of two independent Gaussian noise sources  \cite{ho2020denoising,luo2022understanding}, $x_t$ can be directly obtained by
\begin{equation}
    \label{formula:qxt0}
    q(x_t|x_0)= \mathcal{N}(x_t;\sqrt{\bar{a}_t}x_0,(1-\bar{a}_t)\boldsymbol{I}),
\end{equation}
where $a_t=1-\beta_t$, $\bar{a}_t=\prod^t_{t'=1}a_{t'}$, and $x_t$ can be re-parameterized as $x_t=\sqrt{\bar{a}_t}x_0+\sqrt{1-\bar{a}_t}\epsilon, \epsilon\in\mathcal{N}(\boldsymbol{0},\boldsymbol{I})$. To regulate the added noises in $x_{1:T}$, we designed the following linear noise schedule for $1-\bar{a}_t$:
\begin{equation}
1-\bar{a}_t=s\cdot(a_{min}+\frac{t-1}{T-1}[a_{max}-a_{min}]), t \in \{1,\dots,T\},
\end{equation}
where hyperparameter $s \in[0,1]$ controls the noise scales, and $a_{min}<a_{max}\in(0,1)$ indicates the upper and lower bounds of the added noise.

\mypara{Reverse Process.} Starting from $X_T$, the reverse process gradually recovers users' features by denoising the transition step:
\begin{equation}
    p_{\theta}(x_{t-1}|x_t)=\mathcal{N}(x_{t-1};\mu_\theta(x_t,t),\Sigma_\theta(x_t,t)),
\end{equation}
where $\mu(x_t,t)$ and $\Sigma_\theta(x_t,t)$ are the Gaussian parameters outputted by any neural network with trainable parameters $\theta$.
\subsubsection{Training} To learn $\theta$, \DiffBot aims to maximize the Evidence Lower Bound (ELBO) of the observed user features $x_0$:
\begin{equation}
\label{formula:ELBO}
\begin{aligned}
    \log p(x_0) &=  \log \int p(x_{0:T})\text{d}x_{1:T} = \log \mathbb{E}_{q(x_{1:T}|x_0)}\left[ \frac{p(x_{0:T})}{q(x_{1:T}|x_0)}\right] \\
    &\underbrace{\ge \mathbb{E}_{q(x_1|x_0)}[\log p_\theta(x_0|x_1)]}_{\text{(reconstruction term)}} - \underbrace{D_{KL}(q(x_T|x_0)\parallel p(x_T))}_{\text{(prior matching term)}} \\
    &-\sum^{T}_{t=2} \underbrace{\mathbb{E}_{q(x_t|x_0)}[D_{KL}(q(x_{t-1}|x_t,x_0)\parallel p_\theta(x_{t-1}|x_t))]}_{\text{(denoising matching term)}},
\end{aligned}
\end{equation}
where 1) the prior matching term is a constant without trainable parameters and thus can be ignored during training;
2) the reconstruction term denotes the recovery probability of $x_0$ while the denoising matching terms regulate the recovery of $x_{t-1}$, where $t$ varies from 2 to $T$ in the reverse process; 
3) the denoising matching terms regulate $p_\theta(x_{t-1}|x_t)$ to align with the tractable ground-truth transition step $q(x_{t-1}|x_t,x_0)$ via KL divergence to optimize $\theta$ by iteratively recovering $x_{t-1}$ from $x_t$. 
The denoising matching terms can be simplified as $\sum^T_{t=2}\mathbb{E}_{t,\epsilon}[\parallel\epsilon-\epsilon_\theta(x_t,t)\parallel^2_2]$ \cite{ho2020denoising}, $\epsilon\sim\mathcal{N}(0,\boldsymbol{I})$; and $\epsilon_\theta$ is parameterized by a neural network to predict the noises $\epsilon$ that determines $x_t$ from $x_0$ in the forward process \cite{luo2022understanding}. 
In summary, the optimization goal of training is to maximize the reconstruction term and the denoising matching terms.

\mypara{Estimation of the denoising matching terms.} 
Using Bayes' rules, $q(x_{t-1}|x_t,x_0)$ can be rewritten as the following closed-form \cite{luo2022understanding}:
\begin{equation}
\begin{aligned}
q(x_{t-1}|x_t,x_0) \propto \mathcal{N}(x_{t-1};\tilde{\boldsymbol{\mu}}(x_t,x_0,t),\sigma^2(t)\boldsymbol{I}), \text{where}\\
\left\{
\begin{aligned}
\label{formula:mubar}
\tilde{\boldsymbol{\mu}}(x_t,x_0,t)&= \frac{\sqrt{a_t}(1-\bar{a}_{t-1})}{1-\bar{a}_t}x_t + \frac{\sqrt{\bar{a}_{t-1}}(1-a_t)}{1-\bar{a}_t}x_0, \\
\sigma^2(t)&=\frac{(1-a_t)(1-\bar{a}_{t-1})}{1-\bar{a}_t}.
\end{aligned}
\right.
\end{aligned}
\end{equation}
Here, $\tilde{\boldsymbol{\mu}}(x_t,x_0,t)$ and $\sigma^2(t)\boldsymbol{I}$ are the mean and covariance of $q(x_{t-1}|x_t,x_0)$. To keep training stable and simplify the calculation, we do not learn $\Sigma_\theta(x_t,t)$ in $p_\theta(x_{t-1}|x_t)$ and instead directly set $\Sigma_\theta(x_t,t)=\sigma^2(t)\boldsymbol{I}$ following \cite{ho2020denoising}. 
Then, the denoising matching term $\mathcal{L}_t$ at step $t$ can be calculated by
\begin{equation}
\label{formula:lt}
\begin{aligned}
    \mathcal{L}_t &\triangleq \mathbb{E}_{q(x_t|x_0)}[D_{KL}(q(x_{t-1}|x_t,x_0)\parallel p_\theta(x_{t-1}|x_t))] \\
    &=\mathbb{E}_{q(x_t|x_0)}\left[\frac{1}{2\sigma^2(t)}[\parallel\boldsymbol{\mu}_\theta(x_t,t)-\bar{\boldsymbol{\mu}}(x_t,x_0,t)\parallel^2_2] \right],
\end{aligned}
\end{equation}
which pushes $\boldsymbol{\mu}_\theta(x_t,t)$ toward $\bar{\boldsymbol{\mu}}(x_t,x_0,t)$. Following Eq.~\ref{formula:mubar}, we can factorize $\boldsymbol{\mu}_\theta(x_t,t)$ as
\begin{equation}
\label{formula:mutheta}
\boldsymbol{\mu}_\theta(x_t,t)= \frac{\sqrt{a_t}(1-\bar{a}_{t-1})}{1-\bar{a}_t}x_t + \frac{\sqrt{\bar{a}_{t-1}}(1-a_t)}{1-\bar{a}_t}\hat{x}_\theta(x_t,t),
\end{equation}
where $\hat{x}_\theta(x_t,t)$ is the predicted $x_0$ based on $x_t$ and $t$. We bring Eq.~\ref{formula:mubar} and Eq.~\ref{formula:mutheta} into Eq.~\ref{formula:lt} to regulate $\hat{x}_\theta(x_t,t)$ to predict $x_0$ accurately: 
\begin{equation}
\label{formula:Lt}
    \mathcal{L}_t = \mathbb{E}_{q(x_t|x_0)} \left[ \frac{1}{2} \left(\frac{\bar{a}_{t-1}}{1-\bar{a}_{t-1}} - \frac{\bar{a}_t}{1-\bar{a}_t} \right ) \parallel \hat{x}_\theta(x_t,t)-x_0 \parallel^2_2  \right ]
\end{equation}
In short, we use a neural network (an MLP in this work) to implement $\hat{x}_\theta(x_t,t)$ and calculate Eq.~\ref{formula:Lt} to estimate the denoising matching terms. Moreover, $\hat{x}_\theta(\cdot)$ with $x_t$ and the step embedding of $t$ as input can be used to predict $x_0$.

\mypara{Estimation of the reconstruction term.} We define $\mathcal{L}_1$ as the negative of the reconstruction term in Eq.~\ref{formula:ELBO} and calculate $\mathcal{L}_1$ by 
\begin{equation}
\label{formula:l1}
\begin{aligned}
    \mathcal{L}_1&\triangleq - \mathbb{E}_{q(x_1|x_0)}[\log p_\theta(x_0|x_1)] \\
    &=\mathbb{E}_{q(x_1|x_0)}[\parallel \hat{x}_\theta(x_1,1)-x_0\parallel^2_2],
\end{aligned}
\end{equation}
where we use the unweighted $-\parallel \hat{x}_\theta(x_1,1)-x_0\parallel^2_2$ to estimate the Gaussian log-likelihood $\log p(x_0|x_1)$.

\mypara{Optimization.} 
Using Eq.~\ref{formula:Lt} and Eq.~\ref{formula:l1}, the Eq.~\ref{formula:ELBO} can be formulated as $-\mathcal{L}_1-\sum^T_{t=2}\mathcal{L}_t$. 
To maximize the ELBO, we can optimize $\theta$ in $\hat{x}_\theta(x_t,t)$ by minimizing $\sum^T_{t=1}\mathcal{L}_t$. We uniformly sample step $t$ to optimize the expectation $\mathcal{L}(x_0,\theta)$ over $t\sim\mathcal{U}(1,T)$ as follows:
\begin{equation}
    \mathcal{L}(x_0,\theta) = \mathbb{E}_{t\sim\mathcal{U}(1,T)}\mathcal{L}_t.
\end{equation}
The training procedure of \DiffBot is presented in Alg. \ref{algorithm1}.

Alg.~\ref{algorithm1} summarizes the training process of \DiffBot. 
First, batches of user features are repeatedly sampled from $X$. Then, a time step $t$ is sampled, and the selected loss function is computed based on $t$ (lines \ref{alg1:sample2}-\ref{alg1:loss_fuc}). 
Finally, gradient backpropagation is used to update the optimization parameter(line \ref{alg1:backpropagation}). The process is repeated until the parameters converge.

\subsubsection{Bot Generation} 
We used \DiffBot to model an attacker's replication of other account data to evade detection.
Generating user features from random Gaussian noise would compromise the user's personalized information. 
Therefore, we propose a simple bot account generation strategy. 
\DiffBot first corrupts $x_0$ by $x_0\to x_1\to \cdots \to x_{T'}$ for $T'$ steps in the forward process, and then sets $\hat{x}_{T}=x_{T'}$ to execute reverse denoising $\hat{x}_T\to\hat{x}_{T-1}\to\cdots\to\hat{x}_0$ for $T$ steps. 
The reverse denoising ignores the variance and uses $\hat{x}_t=\mu_\theta(\hat{x}_t,t)$ via Eq.\ref{formula:mutheta} to generate bots. The generation procedure is summarized in Alg.\ref{algorithm2}. 
The generated user set $U_g$ is controlled by the RL agent as the evolving bots.

Alg.~\ref{algorithm2} describes the bot generation process using the \DiffBot model. 
First, a user $x_u$ is selected to replicate (Line \ref{alg2:sample}), and its feature is set to be the initial user feature for the forward propagation of the diffusion model (Line \ref{alg2:setx0}). 
The final noisy feature $x_{T'}$ is directly computed based on the predetermined final time step $T'$, as specified by Eq.~\ref{formula:qxt0} (line \ref{alg2:XT}). 
Subsequently, the noisy feature $x_{T'}$ is repeatedly denoised to generate replicated vectors $\hat{x}_0$ (lines \ref{alg2:denoise}-\ref{alg2:denoise_end}), which are added to the generated user set $U_g$ (line \ref{alg2:add_user}). 
This process is iterated until a predetermined number of generated users is reached.

\begin{algorithm}[t]
\caption{\DiffBot Training.}
\label{algorithm1}
\KwIn{All labelled user features $X$ and randomly initialized $\theta$.}
\Repeat{$\theta$ \rm{converged}}{
    Sample a batch of user features $X_b\subset X$ \label{alg1:sample}\;
    \For {\rm{all} $x_0 \in X_b$ }{ \label{alg1:sample2}
    Sample $t\sim \mathcal{U}(1,T)$ or $t\sim p_t,\epsilon \sim \mathcal{N}(0,\boldsymbol{I})$ \;
    Compute $x_t$ given $x_0,t$ and $\epsilon$ via $q(x_t|x_0)$ in Eq.\ref{formula:qxt0} \;
    Compute $\mathcal{L}_t$ by Eq.\ref{formula:Lt} if $t>1$, otherwise by Eq.\ref{formula:l1} \; \label{alg1:loss_fuc}
    Calculate $\nabla_{\theta}\mathcal{L}_t$ to optimize $\theta$ \; \label{alg1:backpropagation}
    }
} 
\KwOut{Optimized $\theta$}
\end{algorithm}

\begin{algorithm}[t]
\caption{Bot Generation.}
\label{algorithm2}
\KwIn{$\theta$, the sampled user feature set $U_b$, generated user set $U_g$}
\For {$u\in\mathcal{U}_b$}{  \label{alg2:sample} 
    Set $x_0=x_u$ and sample $\epsilon \sim \mathcal{N}(0,\boldsymbol{I})$ \; \label{alg2:setx0}
    Compute $x_{T'}$ via $x_0,T',\epsilon$ via Eq.\ref{formula:qxt0}, and set $\hat{x}_T=x_{T'}$\; \label{alg2:XT}
    \For {$t=T,\cdots$,1}{ \label{alg2:denoise}
        $\hat{x}_{t-1}=\mu_\theta(\hat{x}_t,t)$ calculated from $\hat{x}_t$ and $\hat{x}_\theta(\cdot)$ via Eq.\ref{formula:mutheta} \;
    } \label{alg2:denoise_end}
    Add $\hat{x}_0$ to $U_g$\; \label{alg2:add_user}
} 
\KwOut{Generated $U_g$}
\end{algorithm}

\subsection{Social Bots Control Framework}\label{sec:robctrl}

\begin{figure}[t]
\centering
\includegraphics[width=0.8\textwidth]{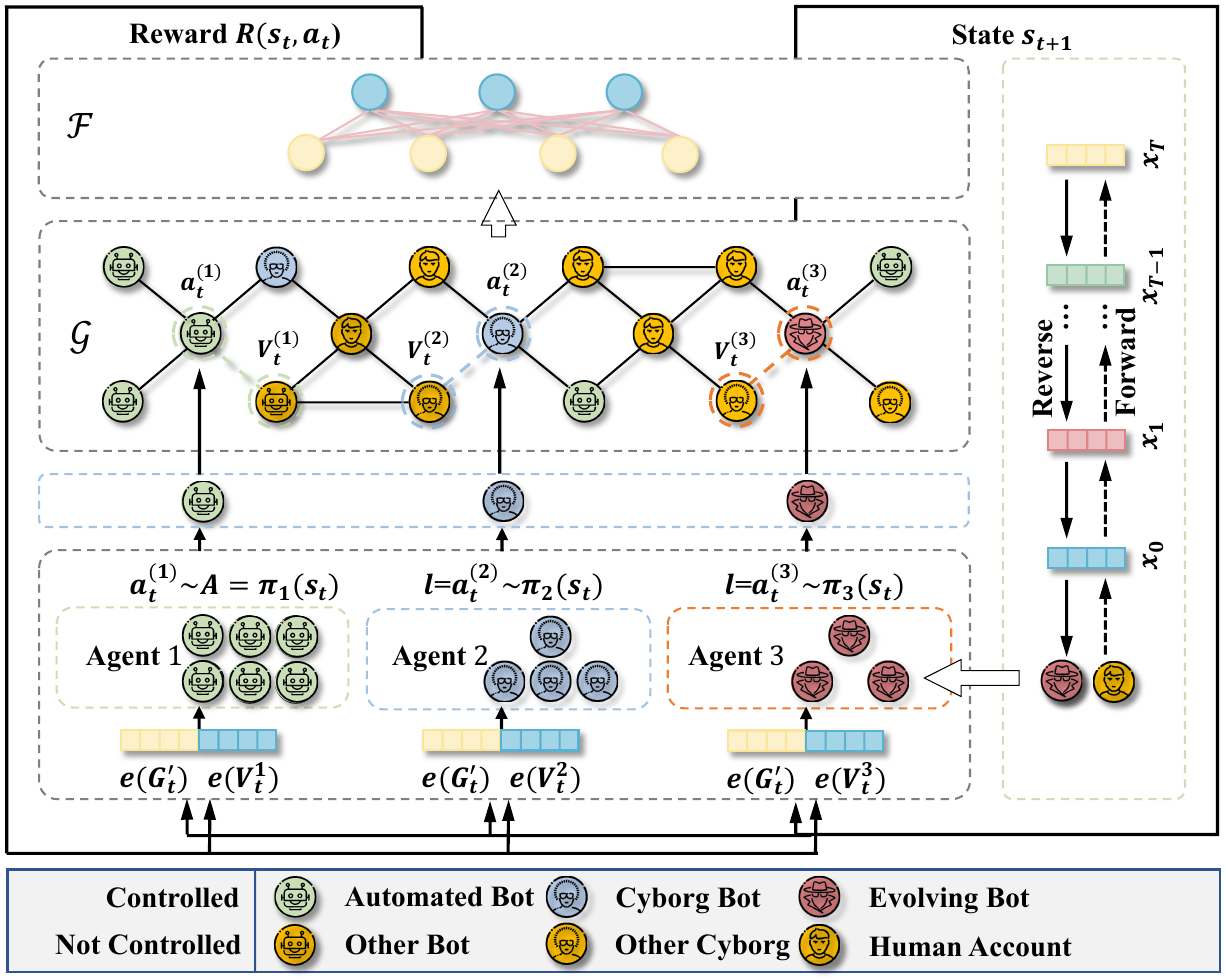}
\caption{The proposed \RoBCtrl framework.}
\label{fig:framework}
\end{figure}

We use an agent to control each generated bot individually. These bots---including evolving bots, automated bots, and cyborg bots as introduced in \S~\ref{sec:3.2}---are used to outcomes the results of the social bot detector. 
To model the malicious behavior of different populations of bot accounts, we use MARL to simulate the collaborative actions of different populations attacking detectors with different agents. 
Figure \ref{fig:framework} illustrates the proposed multi-agent \textbf{\RoBCtrl} framework. 
First, we create a group of automated bots and a group of cyborg bots. 
Next, we generate a group of evolving bots using \DiffBot based on benign accounts selected from existing users. Each agent then selects a target account and takes an action. 
These actions are then applied to the social network graph within the surrogate detector's constructed environment. 
The updated state and rewards are subsequently returned to each agent for optimization.

\subsubsection{Multi-agent reinforcement learning} 
\label{method:marl}
The MARL process is a Markov game. 
We make the following formalization.

\begin{definition}[Markov Game]
A Markov game can be defined by $(\mathbb{N},\mathcal{S},\{\mathcal{A}^i\}_{i\in\mathcal{N}},\mathcal{P},\{R^i\}_{i\in\mathbb{N}},\gamma)$, where $\mathbb{N}=\{1,\cdots, N\}$ represents the agent set, $\mathcal{S}$ is the state space that can be observed by all agents, $\mathcal{A}^i$ is the action space of agent $i$, $\mathcal{P}$ is the state transaction under a state $s\in\mathcal{S}$ and action $a\in\mathcal{A}$, $R$ is the reward function, and $\gamma$ is the discount factor for long term reward.
\end{definition}
\mypara{State.} All agents work cooperatively to attack the same target accounts against the detector $f$. They share the same state at time $t$ represented by concatenated vector $[\mathrm{pool}(e(\mathcal{G}_t)), e(\mathcal{G}_t)_{[i]}]$ where $\mathcal{G}_t$ is the perturbed graph at time $t$, $e$ is the graph encoder (e.g., GCN, SAGE), and $i$ represents the $i$-th target account.

\mypara{Action.}
Each agent controls a group of bot accounts $U^i_c$, and each $u\in U^i_c$ adds edges to the target account $v\in U_T$.
We use $a^i_t(u,v)$ to represent the action that adds the edge from controlled bot $u$ to the target account $v$ in timestep $t$. 
The action space of agent $i$ at time $t$ is $a^i_t\in\mathcal{A}^i\subseteq U_c^i \times U_T$. 

\mypara{Reward.}
In the black-box attack, we first select a classifier and aim to flip its classification results. We use a perturbed graph to attack the detector, whose architecture is unknown.
Due to the lack of knowledge about the detector's GNN architecture, we default to using a classic GNN model as our surrogate detector and use its classification results on the target user $U_T$ as the reward value to guide the agent. 
When all agents take actions according to the budget, the reward value for each agent towards the target account $v\in U_T$ is given as follows:
\begin{equation}
    \label{eq:reward}
    R((\mathcal{G}_t,f_{\theta^*})_v)=\left\{
    \begin{aligned}
    1:f_{\theta ^*}(\mathcal{E}_t,X)_v &\neq  y_v,\\ -1:f_{\theta ^*}(\mathcal{E}_t,X)_v & = y_v.
    \end{aligned}
    \right.
\end{equation}

\mypara{Terminal.} The Markov process stops after each agent has modified the graph according to the number budget of edges added $\Delta^i_e$, as specified by the budget. 

\subsubsection{Deep Q-learning} 
To solve the Markov game, we must determine the optimal policy that maximizes the expected value of long-term rewards.
Each agent has its controlled user accounts and a budget for the number of edges to be added. 
Each proxy $i$ should have its own policy $\pi^i$ to take action $a_t^i\sim\pi^i(\cdot|s_t)$. 
We use Q-learning to learn the optimal policy $\pi^{i,*}$ parameterized by the Q-function of the neural network $Q^{i,*}(s_t,a_t)$. 
The optimal Q-value for an agent $i$ can be represented by the following Bellman equation:
\begin{equation}
    \label{eq:rl_optimize}
    Q^{i,*}(s_t,a_t^i)=R(s_t,a_t^i)+\gamma \max_{a'}Q^*(s_{t+1},a^{i,'}),
\end{equation}
where the $a^{i,'}$ represent the future actions based on the state $s_t$. 
When attacking the detector, we will use a greedy policy to determine the best action for agent $i$ based on $s_t$ through maximizing the $Q$-value as
\begin{equation}
\label{eq:PI}
    \pi^i(a^i_t|s_t;Q^{i,*})=\mathop{\arg\max}_{a_t^i}Q^{i,*}(s_t,a_t^i).
\end{equation}
For each target bot $v\in U_T$, the agent chooses the controlled user $u\in U_c$ with the most influence on $v$ to flip the GNN detector result for $v$. 
We use a two-layer GCN to parameterize the Q-function to obtain the embedding of target node $h_{v,t}$ and perturbed graph $\mathcal{G}_t$ as the current state $s_t$.
For agent $i$ at time $t$ with embeddings of controlled user accounts $h_{u,t},u\in U_c^i$, the target bot node $h_{v,t},v\in V_T$, and graph $h_{\mathcal{G}_t}$, the Q-value of action $a_t^i(u,v)$ is calculated as follows:
\begin{equation}
Q^i(s_t,a^i_t(u,v))=\sigma \left (W^1([h_{u,t},h_{\mathcal{G}_t}]) \right)\sigma \left(W^2(h_{v,t}) \right),
\end{equation}
where $W^1$ and $W^2$ represent two linear layers applied on the target embeddings and concatenated embeddings $[h_{u,t},h_{\mathcal{G}_t}]$  to obtain the Q-value of the given action. The attack process is summarized in Alg. \ref{algorithm3}.

Alg.~\ref{algorithm3} describes the attack process of \RoBCtrl on a GNN detector for social bot attacks.
First, the parameters of the \DiffBot model are trained (line \ref{alg3:traindiff}), and then the model is used to generate evolved bots (line \ref{alg3:genbots}). 
The nodes and features of these evolved bots are added to graph $\mathcal{G}$ (line \ref{alg3:addbots}). 
Next, the MARL parameters are initialized (line \ref{alg3:initialagents}). 
Using the $i$-th identified node (line \ref{alg3:choosev}), we calculate the environmental state value of the node (Line \ref{alg3:calstate}). 
Reinforcement learning agents select an action (lines \ref{alg3:calaction}-\ref{alg3:chooseaction}), and the edge corresponding to that action is added to the poisoned graph $\mathcal{G}_t'$ (line \ref{alg3:addedge}). 
We then compute the reward value (Line \ref{alg3:calreward}), store the state transition entries into memory (Line \ref{alg3:storememory}), and optimize the parameters of the agents' policy networks (Line \ref{alg3:optiagent}). 
This process is repeated until the parameters of the reinforcement learning agents converge.
Finally, the poisoned graph $\mathcal{G}'$ is obtained by using the trained reinforcement learning agent to attack graph $\mathcal{G}$ (line \ref{alg3:attack}).

\begin{algorithm}[t]
\caption{\RoBCtrl.}
\label{algorithm3}
\KwIn{Social graph $\mathcal{G}$, pre-trained surrogate model $f_{\theta}*$, action aggregation probability $\mathcal{P}$, memories $\mathcal{M}$}
\DiffBot training defined in Alg.\ref{algorithm1}\; \label{alg3:traindiff}
Evolved bot generation defined in Alg.\ref{algorithm2}\; \label{alg3:genbots}
Add the evolved bots' features to $\mathcal{G}$ \; \label{alg3:addbots}
Initialize $N$ RL agents \;  \label{alg3:initialagents}
\For{$t=0,1,\ldots,T$}{
    Query the index $i$ of $v_t$\; \label{alg3:choosev}
    $s_t=[\mathrm{pool}(e(\mathcal{G}_t)),e(\mathcal{G}_t)_{[i]}]$ \; \label{alg3:calstate}
    $a_t'=(a_t^1,a_t^2,a_t^3)$ via Eq.\ref{eq:PI}\;  \label{alg3:calaction}
    Choose action $a_t$ from $a_t'$ according $\mathcal{P}$ \;   \label{alg3:chooseaction}
    Add edge $(i,a_t)$ to obtain $\mathcal{G}_t'$\; \label{alg3:addedge}
    Obtain $R((\mathcal{G}_t,f_{\theta*})_v)$ via Eq.\ref{eq:reward} \; \label{alg3:calreward}
    \For{$j=1,2,\ldots,N$}{
        Store the $T^j_t=(s_t,a^j_t,s_{t+1}, R((\mathcal{G}_t,f_{\theta*})_v))$ into $\mathcal{M}_j$ \; \label{alg3:storememory}
    }
    \For{$j=1,2,\ldots,N$}{
        Optimize $\pi^j$ via Eq.\ref{eq:rl_optimize}\; \label{alg3:optiagent}
    }
}
Attack $\mathcal{G}$ via agents to obtain $\mathcal{G}'$\;  \label{alg3:attack}
\KwOut{Perturbed Social Graph $\mathcal{G}'$}
\end{algorithm}

\subsubsection{State Abstraction}
\label{section: stateabstraction}

\begin{figure}[t]
\centering
\includegraphics[width=0.8\textwidth]{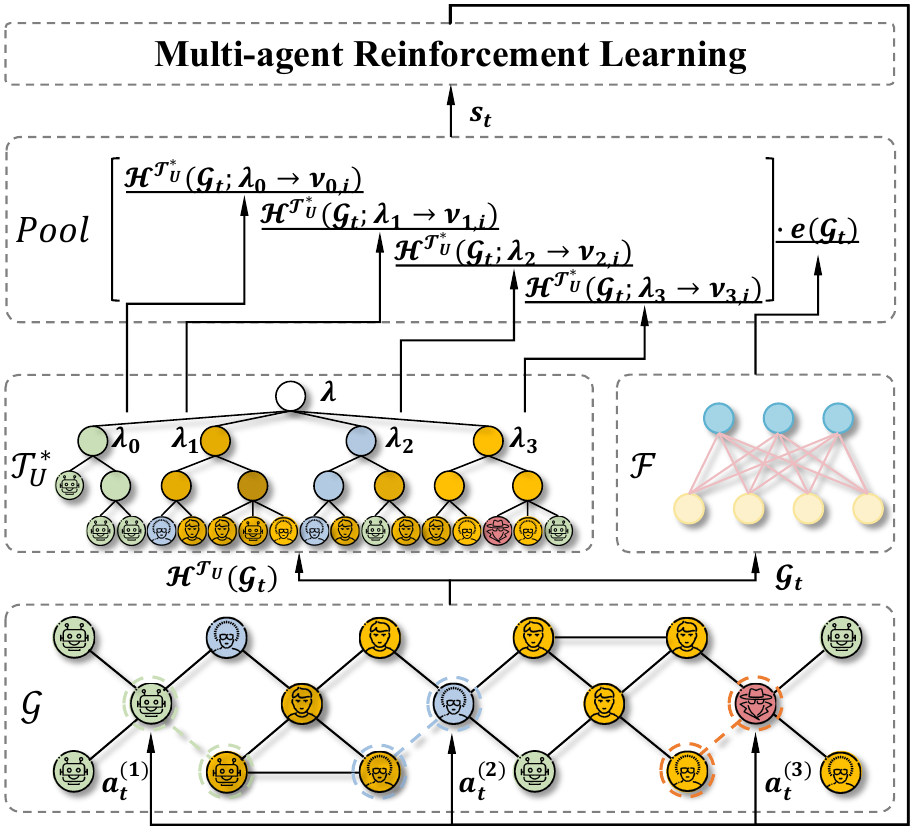}
\caption{Proposed \RoBCtrl framework optimized by hierarchical state abstraction.}
\vspace{-0.5cm}
\label{fig: robctrl_sa}
\end{figure}

To further enhance attack performance and accelerate the attack process, we introduce a hierarchical state abstraction mechanism \cite{zeng2023hierarchical} into \RoBCtrl. 
This mechanism replaces the original average pooling operation (i.e., $\mathrm{pool}(e(\mathcal{G}))$ defined in \S~\ref{method:marl}), providing a more refined state representation. 
As illustrated in Fig. \ref{fig: robctrl_sa}, it is positioned between the homogeneous social network graph $\mathcal{G}$ and the MARL process.

The state abstraction method proposed by \cite{zeng2023hierarchical} is based on minimizing structural entropy~\cite{li2016structural}, a metric to quantify the uncertainty within the graph structure.
Given the graph $\mathcal{G}=(V,E)$ and its corresponding encoding tree $\mathcal{T}$, the structural entropy of non-root node $v_\tau$ can be calculated by
\begin{equation}
    \mathcal{H}^\mathcal{T}(\mathcal{G};v_\tau) = -\frac{g_{v_\tau}}{vol(V)}\log\frac{vol(v_\tau)}{vol(v_\tau^+)},
\end{equation}
where $v_\tau$ represents a node in $\mathcal{T}$ (excluding the root node $\lambda$) and corresponds to a subset $V_\tau \subseteq V$. 
Here, $g_{v_\tau}$ denotes the number of edges connecting nodes inside $\mathcal{V}_\tau$ with those outside, $v_\tau^+$ is the immediate predecessor of $v_\tau$, and $vol(v_\tau)$, $vol(v_\tau^+)$, and $vol(V)$ refer to the sum of the degrees of nodes in $v_\tau$, $v_\tau^+$, and $V$, respectively. 
The structural entropy of graph $\mathcal{G}$ is the minimum structural entropy across all possible encoding trees: $\mathcal{H}^{\mathcal{T}^*}(\mathcal{G}) = \min_{\forall \mathcal{T}} \{\mathcal{H}^\mathcal{T}(\mathcal{G})\}$. 
The optimal encoding tree $\mathcal{T}^*$ naturally represents a multi-grained hierarchical aggregation and division. 
Based on this observation, we use structural entropy as a tool to decode the hierarchical structure of a graph into an encoding tree, which can be used to filter out irrelevant environmental information and compress the original state space.

At regular intervals $t_{up}$, we minimize the structural entropy of the perturbed graph $\mathcal{G}_t$ to derive a tree structure partitioning $\mathcal{T}^*_U$ for all accounts.
In this structure, each tree node $v \in \mathcal{T}^*_U$ corresponds to a subset of accounts, and each leaf node $\nu$ corresponds to a subset containing only one account.
The subsets of the root node $\lambda$'s children are designated as account communities in our work.
Moving from left to right in $\mathcal{T}^*_U$, the child nodes of the root are marked as $\lambda_i$, where $i$ increases from $0$.
Within each community $U_i$ of node $\lambda_i$, we employ the path entropy \cite{zeng2023hierarchical} to calculate the pooling matrix $P$ as follows:
\begin{equation} \label{equ: matrix p}
    P_{i,j} = \frac{\mathcal{H}^{\mathcal{T}^*_U}(\mathcal{G}_t;\lambda_i \rightarrow \nu_{i,j})}{\sum_{k} \mathcal{H}^{\mathcal{T}^*_U}(\mathcal{G}_t;\lambda_i \rightarrow \nu_{i,k})}\text{.}
\end{equation}
Here, $\nu_{i,j}$ is a leaf node corresponding to each account $v_j$ in $U_i$.
The path entropy $\mathcal{H}^{\mathcal{T}^*_U}(\mathcal{G}_t;\lambda_i \rightarrow \nu_{i,j})$ is defined as the sum of the structural entropy of nodes along the path from $\lambda_i$ to $\nu_{i,j}$.

On the one hand, the pooling operation induced by matrix $P$ reduces the dimensionality of the network representation from the number of leaves to the number of the root's children in $\mathcal{T}^*_U$, thereby decreasing the complexity of multi-agent exploration.
On the other hand, the presence of disjoint communities results in the matrix $P$ being a sparse block diagonal matrix, which ensures computational efficiency in the pooling operation $P \cdot e(\mathcal{G}_t)$, even in large-scale social networks.

Alg. \ref{alg: attack_sa} summarizes the attack process optimized by state abstraction.
At each step, $t$, \RoBCtrl applies the pooling matrix $P$ to obtain the state representation $s_t$ (line \ref{alg4: line 11}). 
Periodically, on the basis of perturbed graph structure $\mathcal{G}_t$, the partitioning structure $\mathcal{T}^*_U$ is re-derived (line \ref{alg4: line 7}), and the pooling matrix $P$ is re-calculated (line \ref{alg4: line 8}).

\begin{algorithm}[t]
\caption{\RoBCtrl with State Abstraction.}
\label{alg: attack_sa}
\KwIn{Social graph $\mathcal{G}$, pre-trained surrogate model $f_{\theta}*$, action aggregation probability $\mathcal{P}$, memories $\mathcal{M}$, update interval $t_{up}$}
\DiffBot training defined in Alg.\ref{algorithm1}\;
Evolved bot generation defined in Alg.\ref{algorithm2}\; 
Add evolved bots' features to $\mathcal{G}$ \;
Initialize $N$ RL agents \;
\For{$t=0,1,\ldots,T$}{
    \If{$t \mod t_{up} == 0$}
    {
        Derive the optimal encoding tree $\mathcal{T}^*_U$ of the perturbed graph $\mathcal{G}_t$\; \label{alg4: line 7}
        Update the pooling matrix $P$ via Eq. \ref{equ: matrix p}\; \label{alg4: line 8}
    }
    Query the index $i$ of $v_t$\;
    $s_t=[P \cdot e(\mathcal{G}_t),e(\mathcal{G}_t)_{[i]}]$ \; \label{alg4: line 11}
    Lines \ref{alg3:calaction} - \ref{alg3:optiagent} in Alg. \ref{algorithm3}
}
Attack $\mathcal{G}$ via agents to obtain $\mathcal{G}'$\;
\KwOut{Perturbed Social Graph $\mathcal{G}'$}
\end{algorithm}

\section{EVALUATION}
\label{sec:Evaluation}
Our experiments were designed around a black-box scenario to emulate the process of bots interacting with other users while evading detection without any knowledge of the underlying model architectures. 
Specifically, the experiments aimed to answer the following Research Questions (RQs):
\begin{itemize}[leftmargin=*]
    \item \textbf{RQ1}. How effective is \RoBCtrl in attacking GNN-based detectors when trained on normal-scale or large-scale social graphs?
    \item \textbf{RQ2}. Does the poisoned graph retain its key statistics after an attack by \RoBCtrl ?
    \item \textbf{RQ3}. What attack strategies does \RoBCtrl simulate, and are there any countermeasures against these attacks?
    \item \textbf{RQ4}. To what extent does the state abstraction mechanism enhance the attack performance of the \RoBCtrl framework?
\end{itemize}
\begin{table}
    \caption{Statistics of datasets.}
    \label{tab:statistics}
    \centering
    \scalebox{1}{
    \begin{tabular}{ccccc}
    \hline
    \toprule
    \specialrule{0em}{1pt}{1pt}
        \textbf{Dataset}                           &    \textbf{Nodes}  &  \textbf{Edges}    &  \textbf{Human}    & \textbf{Bot}   \\
    \specialrule{0em}{1pt}{1pt}
    \hline
    \specialrule{0em}{1pt}{1pt}
        \textbf{Twibot-20} \cite{feng2021twibot}    &  229580   &  227979   &   5237    &  6589 \\
        \textbf{MGTAB-Large} \cite{shi2023mgtab}    &  410199   &  233577   &    7451   &  2748    \\
        \textbf{Cresci-15} \cite{cresci2015fame}    &   5301    &  14220    &    1950   &  3351 \\
       \textbf{MGTAB}  \cite{shi2023mgtab}         &   10199   &  1700108  &    7451   &  2748  \\
    \bottomrule
    \hline
    \end{tabular}}
\end{table}

\subsection{Experimental Setup}

\subsubsection{Software and Hardware} 

\RoBCtrl was implemented using Python 3.9.16 and PyTorch 1.13.1 and runs on two servers; one is equipped with NVIDIA Tesla V100 GPU, 2.20GHz Intel Xeon Gold 5220 CPU, and 512GB RAM, and the other is equipped with NVIDIA Tesla A100 GPU, 2.25GHz AMD EPYC 7742 CPU, and 1024GB RAM.
\subsubsection{Datasets}
To evaluate \RoBCtrl on large-scale graph structures, we choose two datasets \textbf{Twibot-20} \cite{feng2021twibot} and \textbf{MGTAB-Large} \cite{shi2023mgtab} which represent the latest advancements in social bot research. 
\textbf{Twibot-20} encompasses user information from four domains, i.e., politics, business, entertainment, and sports. 
\textbf{MGTAB-Large} offers a high-quality dataset, meticulously annotated by expert-level annotators that serves as a standardized benchmark for graph-based bot detection. 
For normal-scale graph data, we select two datasets \textbf{Cresci-15} \cite{cresci2015fame} and \textbf{MGTAB} \cite{shi2023mgtab}.
\textbf{MGTAB} \cite{shi2023mgtab} is the unsupported user version of the \textbf{MGTAB-Large} \cite{shi2023mgtab} dataset.  
Table \ref{tab:statistics} comprehensively overviews the dataset statistics.


In a black-box setting, attackers lack knowledge of the detector's model architecture, but they are privy to the graph structure data and initial features of the nodes they intend to attack. 
Consequently, the attacker must use a surrogate model $\mathcal{M}$ for the target GNN architecture to be attacked. 
On the basis of $\mathcal{M}$, the attacker attacks the graph structure data and ultimately uses the attacked graph structure to attack the target detector during the platform testing phase. 
In our experiments, we selected a GCN model to serve as the surrogate model. 
To preserve equitability in the challenge of compromising the target model, we incorporate a suite of prevalently adopted GNN architectures (e.g., GCN \cite{kipf2016semi}, GAT \cite{velivckovic2017graph}, GraphSAGE \cite{hamilton2017inductive}, AirGNN \cite{liu2021graph}, APPNP \cite{gasteiger2018predict}, and SGC \cite{wu2019simplifying}) as the deployed detectors or target models.
To align with production settings in practical implementations, $\mathcal{M}$ and the target models can be replaced with state-of-the-art bot detection models.

\subsubsection{Baselines}
Baseline approaches, typically categorized as either \textbf{feature}-based or \textbf{gradient}-based, assume that all interactive data on the platform and any individual user information can be modified. This assumption is \textbf{inconsistent} with real-world social bot detection scenarios and therefore, these approaches were not employed in our investigations.  
Furthermore, the gradient-based baseline approach requires an \textbf{impractically high} amount of GPU memory to process large-scale graph data, making it unsuitable for practical use.
Thus, we selected global poisoning attack methods, including \textbf{Random}, \textbf{DICE} \cite{waniek2018hiding}, and textbf{Topology} \cite{xu2019topology}  as baselines. 
To compare and demonstrate the superiority of our method for attacking normal-scale graph data, we also selected \textbf{FGA} \cite{chen2018fast} and \textbf{IGA} \cite{wu2019adversarial}, two target attack methods, as experimental baselines. 
The \textbf{Random} method randomly adds edges to the graph data until the budget is met. 
The \textbf{DICE} \cite{waniek2018hiding} method employs a heuristic approach to disconnect the target node from its neighboring nodes and subsequently connect it with other neighbors.
This helps the target node avoid detection by the adversary.
The \textbf{Topology} method \cite{xu2019topology} can degrade the classifier's performance by minimizing or maximizing classification loss while only slightly changing the graph structure.
The \textbf{FGA} method \cite{chen2018fast} extracts link gradient information from GCN, and iteratively modifies the graph by selecting the pair of nodes with the highest absolute gradient to attack the target node. 
The \textbf{IGA} method \cite{wu2019adversarial} iteratively selects the edge or feature that exerts the most decisive influence, resulting in a higher rate of target node misclassification by the victim model.

\begin{table*}[t]
\caption{Detection accuracy results after attacks by graph-based methods(\%) in large-scale social bot graph datasets.}
\label{tab:comparision-table-large}
\centering
\scalebox{1}{
\setlength{\tabcolsep}{0.9mm}{
\begin{tabular}{c|cccccc|cccccc}
\hline
\toprule
\specialrule{0em}{1pt}{1pt}
  \multirow{2}{*}{\textbf{Methods}}  &\multicolumn{6}{c|}{\textbf{TwiBot-20}} & \multicolumn{6}{c}{\textbf{MGTAB-Large}} \\ 
\specialrule{0em}{1pt}{1pt}
\cline{2-6}\cline{7-13}
\specialrule{0em}{1pt}{1pt}
     & GCN  & GAT    & SAGE  & AirGNN & APPNP & SGC & GCN  & GAT & SAGE  & AirGNN &  APPNP & SGC \\
\specialrule{0em}{1pt}{1pt}
\hline
\specialrule{0em}{1pt}{1pt}
\textbf{Clean}       & 79.12 & 85.54 & 85.37 & 72.69 & 76.16 & 78.36 & 83.31 & 83.41 & 86.45 & 81.94 & 84.00 & 81.84 \\ 
\specialrule{0em}{1pt}{1pt}
\hline
\specialrule{0em}{1pt}{1pt}
\textbf{Random}      & 78.19 & 85.37 & 83.77 & 72.69 & 74.80 & 78.36 &  83.31 & 83.02 & 84.10 & 80.86 & 82.72 & 81.45\\ 
\textbf{DICE} \cite{waniek2018hiding}    & 77.26 & 84.10 & 84.53 & 68.20 & 73.62  & 77.26 & 78.31 & 77.62 & 84.20 & 76.93 & 78.01 & 77.13\\ 
\textbf{Topology} \cite{xu2019topology}   & 78.61 & 85.29 & 85.03 & 72.86 & 75.57 & 78.78 & / & / & / & / & / & /  \\
\specialrule{0em}{1pt}{1pt}
\hline
\specialrule{0em}{1pt}{1pt}
\textbf{\RoBCtrl-1}        & 73.20 & 59.50 & 84.10 & 67.54 & 75.14 & 73.11 & 72.91 & 73.37 & 85.18 & 73.21 & 72.32 & \textbf{71.83} \\ 
\textbf{\RoBCtrl-2}        & 72.95 & 54.43 & 84.86 & 56.88 & 65.17 & 71.85 & 72.22 & 72.32 & \textbf{83.70} & 72.92 & 72.87 & 72.78\\ 
\textbf{\RoBCtrl}       & \textbf{71.93} & \textbf{54.09} & \textbf{84.01} & \textbf{55.28} & \textbf{65.08} & \textbf{69.06} & \textbf{70.83} & \textbf{72.21}  & 84.59  & \textbf{72.12}  & 7\textbf{2.22} & 72.03 \\ 
\specialrule{0em}{1pt}{1pt}
\hline
\specialrule{0em}{1pt}{1pt}
\textbf{Drop} $\downarrow$ & \textbf{7.19}  & \textbf{31.45} & \textbf{1.36}  & \textbf{17.41}  & \textbf{11.08}  & \textbf{9.30} & \textbf{12.48}  & \textbf{11.20}  & \textbf{2.75} & \textbf{9.82}  & \textbf{11.78} &  \textbf{10.01} \\
\bottomrule
\hline
\end{tabular}}}
\end{table*}

\begin{table*}[tb]
\caption{Detection accuracy results after attacks by graph-based methods(\%) in normal-scale social bot graph datasets.}
\label{tab:normal-scale-comparision-table}
\centering
\scalebox{1}{
\setlength{\tabcolsep}{0.9mm}{
\begin{tabular}{c|cccccc|cccccc}
\hline
\toprule
\specialrule{0em}{1pt}{1pt}
\multirow{2}{*}{\textbf{Methods}} &\multicolumn{6}{c|}{\textbf{Cresci-15}} & \multicolumn{6}{c}{\textbf{MGTAB}} \\ 
\specialrule{0em}{1pt}{1pt}
\cline{2-6}\cline{7-13}
\specialrule{0em}{1pt}{1pt}
    & GCN  & GAT    & SAGE  & AirGNN & APPNP & SGC & GCN  & GAT & SAGE  & AirGNN &  APPNP & SGC \\
\specialrule{0em}{1pt}{1pt}
\hline
\specialrule{0em}{1pt}{1pt}
\textbf{Clean}                              & 91.77& 91.96 & 95.14 & 94.39 & 94.57 & 91.96 & 78.50 & 73.01 & 85.67 & 72.42 & 83.70 & 75.76\\ 
\specialrule{0em}{1pt}{1pt}
\hline
\specialrule{0em}{1pt}{1pt}
\textbf{Random}                             & 89.15 & 89.34 & 88.97 & 90.46 & 91.02 & 90.28 & 78.21 & 72.42 & 85.08 & 72.42 & 83.90 & 75.17\\ 
\textbf{DICE} \cite{waniek2018hiding}       & 87.66 & 87.85 & 91.02 & 90.09 & 89.90 & 89.71 & 77.72 & 72.81 & 84.69 & 72.32 & 83.31 & 74.87 \\ 
\textbf{Topology} \cite{xu2019topology}     & 91.02 & 91.77 & 94.39 & 94.57 & 94.39 & 92.14 & 78.31 & 72.91 & 83.70 & 72.42 & 83.90 & 75.07 \\
\specialrule{0em}{1pt}{1pt}
\hline
\specialrule{0em}{1pt}{1pt}
\textbf{FGA} \cite{chen2018fast}            & 90.65 & 91.77 & 93.83 & 94.20 & 94.01 & 91.96 & 77.91 & 72.42 & 84.78 & 72.32 & 83.70 & 75.26\\
\textbf{IGA} \cite{wu2019adversarial}       & 90.65 & 91.96 & 94.39 & 93.83 & 94.39 & 91.96 & 78.70 & 72.81 & 83.90 & 72.29 & 83.90 & 74.97\\
\specialrule{0em}{1pt}{1pt}
\hline
\specialrule{0em}{1pt}{1pt}
\textbf{\RoBCtrl-1}                         & 74.01 & 57.85 & 59.15 & 47.0  & 52.24 & 59.15 & \textbf{75.26} & 72.42 & 84.49 & 72.29 & 81.84 & 75.26   \\ 
\textbf{\RoBCtrl-2}                         & 73.08 & 53.64 & \textbf{55.23} & \textbf{44.67} & 50.09 & 57.38 & 76.15 & 72.35 & 84.78 & 72.25 & 83.31 & 75.26 \\ 
\textbf{\RoBCtrl}                           & \textbf{63.17} & \textbf{48.22} & 56.63 & 45.23 & \textbf{49.34} & \textbf{53.27} & 76.05 & \textbf{72.22} & \textbf{83.39} & \textbf{72.22} & \textbf{77.13} & \textbf{74.09}\\ 
\specialrule{0em}{1pt}{1pt}
\hline
\specialrule{0em}{1pt}{1pt}
\textbf{Drop $\downarrow$} & \textbf{28.60} & \textbf{43.74} & \textbf{39.91} & \textbf{49.72} &  \textbf{45.23} & \textbf{38.69} & \textbf{3.24} & \textbf{0.79} & \textbf{2.28} & \textbf{0.20} & \textbf{6.57} & \textbf{1.67}\\

\bottomrule
\hline
\end{tabular}}}
\end{table*}

\subsubsection{Implementation and Metrics}
We randomly sampled 100 automated bots and 50 cyborg bots, and we applied the trained diffusion model \DiffBot to create 50 evolved bots for all datasets.
We used PyG \cite{Fey/Lenssen/2019} to implement all GNN detection methods. 
In addition, we use the DeepRobust \cite{li2020deeprobust} to implement the attack baseline methods.
In our approach, we leveraged two dual-layer fully connected networks as the forward network and the backward denoising network of the diffusion model in \DiffBot. 
To further enhance the performance of the RL agents in \RoBCtrl's, we employed a single-layer GCN to aggregate the feature vectors for the nodes after a new-edge insertion. 
Specifically, the feature vector obtained by pooling the entire graph is concatenated with the feature vector of the control node, and the resulting vector is multiplied by the feature vector of the target node.
This product value is then used to select the control node for connection with the target node. 
For more detailed insights, please refer to our \textbf{ code}\footnote{\url{https://anonymous.4open.science/r/RoBCtrl-C35B/}}. 
Given that the datasets of different categories are well balanced and that various attack methods require a significant amount of time to attack large-scale graphs, particularly gradient-based attacks, we randomly selected a random seed and used accuracy as our primary metric; other metrics will also experience a similar decline.

\subsection{RQ1: Attack Performance Comparison}
In response to \textbf{RQ1}, we evaluated how node classification accuracy degraded on both the poisoned large-scale social graph and the normal-scale social graph. 
A greater performance decrease on the poisoned graph indicates more effective attacks. We also conducted ablation experiments on our model to evaluate the effectiveness of different bot type combinations.

\begin{sidewaystable}[thp]
    \setlength{\belowcaptionskip}{16.cm}
    \aboverulesep=0ex
    \belowrulesep=0ex
    \centering
    \renewcommand\arraystretch{1.3}
\caption{Statistics of the clean graph(r=0.00) and the graphs poisoned by \RoBCtrl averaged over five runs, Accuracy (ACC), Average Degree (AD), Largest Connected Component (LCC), Clustering Coefficient (CC), Characteristic Path Length (CPL), Gini Coefficient (GC), Power Law Exp (PLE), and Distribution Entropy (DE).}
\label{tab:statistics-table}
\centering
\scalebox{0.94}{
\setlength{\tabcolsep}{0.5mm}{
\begin{tabular}{ccccccccccccc}
\hline
\toprule
\specialrule{0em}{1pt}{1pt}
 \textbf{Dataset}  & \textbf{$r$} & \textbf{ACC}  & \textbf{AD} & \textbf{LCC} &  \textbf{CC} & \textbf{CPL} & \textbf{GC} & \textbf{PLE} & \textbf{DE} \\ 
\specialrule{0em}{1pt}{1pt}
\hline
\specialrule{0em}{1pt}{1pt}
\multirow{4}{*}{\textbf{Cresci-15}} & 0.0000 & 0.917$\pm$ 0.000 & 1.6132 $\pm$ 0.0000 & 1275.0 $\pm$ 0.0 &  0.0016 $\pm$ 0.0000 & 3.7385 $\pm$ 0.0000 & 0.8694 $\pm$ 0.0000 & 2.3266 $\pm$ 0.0000 & 0.7704 $\pm$ 0.0000 \\ 
& 0.1500 & 0.654 $\pm$ 0.086& 1.7263 $\pm$ 0.0061  & 1778.0 $\pm$ 4.6 & 0.0016 $\pm$ 0.0000 & 3.7942 $\pm$ 0.0351 &  0.8380 $\pm$ 0.0005 & 2.4796 $\pm$ 0.0000 &  0.7917 $\pm$ 0.0003 \\ 
& 0.2000 & 0.514 $\pm$ 0.063 & 1.7711 $\pm$ 0.0034  & 1780.8 $\pm$ 0.4 & 0.0017 $\pm$ 0.0000 &  3.8120 $\pm$ 0.0258 & 0.8325 $\pm$ 0.0003 & 2.3633 $\pm$ 0.0000 & 0.7967 $\pm$ 0.0001 \\ 
& 0.2500 & 0.627 $\pm$ 0.098 & 1.8075 $\pm$ 0.0058  & 1782.6 $\pm$ 1.2 & 0.0017 $\pm$ 0.0000 & 3.7992 $\pm$ 0.0407 & 0.8265 $\pm$ 0.0005 & 2.2620 $\pm$ 0.0000 & 0.8012 $\pm$ 0.0004\\ 
\specialrule{0em}{1pt}{1pt}
\hline
\specialrule{0em}{1pt}{1pt}
\multirow{4}{*}{\textbf{MGTAB}} & 0.0000 &0.785$\pm$ 0.000 & 177.5974 $\pm$ 0.0000 & 10145.0 $\pm$ 0.0 &  0.0008 $\pm$ 0.0000 & 2.2513 $\pm$ 0.0000 & 0.4880 $\pm$ 0.0000 & 1.2121 $\pm$ 0.0000 & 0.9533 $\pm$ 0.0000 \\ 
 & 0.0035 & 0.777 $\pm$ 0.004& 177.5325 $\pm$ 0.2748 & 10151.0 $\pm$ 0.0 & 0.0008 $\pm$ 0.0000 & 2.2482 $\pm$ 0.0006 & 0.4883 $\pm$ 0.0007 & 1.2119 $\pm$ 0.0000 & 0.9531 $\pm$ 0.0001 \\ 
 & 0.0070 &  0.749 $\pm$ 0.031 & 177.4846 $\pm$ 0.1402 & 10152.6 $\pm$ 1.0 & 0.0008 $\pm$ 0.0000 & 2.2470 $\pm$ 0.0013 & 0.4885 $\pm$ 0.0004 & 1.2117 $\pm$ 0.0000 & 0.9529 $\pm$ 0.0001 \\ 
 & 0.0100 & 0.747 $\pm$ 0.022 & 177.8158 $\pm$ 0.0882 & 10153.4 $\pm$ 1.3 & 0.0008 $\pm$ 0.0000 & 2.2448 $\pm$ 0.0013 & 0.4877 $\pm$ 0.0002 & 1.2115 $\pm$ 0.0000 & 0.9529 $\pm$ 0.0001\\ 
 \specialrule{0em}{1pt}{1pt}
\hline
\specialrule{0em}{1pt}{1pt}
\multirow{4}{*}{\textbf{TwiBot}} & 0.0000 & 0.791 $\pm$ 0.000 & 1.9697 $\pm$ 0.0000 & 203279.0 $\pm$ 0.0 & 0.0001 $\pm$ 0.0000 &  / & 0.5626 $\pm$ 0.0000 & 5.6279 $\pm$ 0.0000 & 0.9186 $\pm$ 0.0000\\
& 0.0200 & 0.751 $\pm$ 0.018 & 1.9891 $\pm$ 0.0000& 203484.6 $\pm$ 6.5 & 0.0000 $\pm$ 0.0000& /& 0.5670 $\pm$ 0.0000 & 5.6238 $\pm$ 0.0002 & 0.9156 $\pm$ 0.0000 \\
& 0.0400 & 0.700 $\pm$ 0.027 & 2.0087 $\pm$ 1.7993 & 203478.0 $\pm$ 7.5  & 0.0000 $\pm$ 0.0000 & / & 0.5712 $\pm$ 0.0000 &  5.6215 $\pm$ 0.0002 &  0.9122 $\pm$ 0.0000\\
& 0.0600 & 0.660 $\pm$ 0.043 &  2.0285 $\pm$ 0.0000 & 203482.8 $\pm$ 9.9 & 0.0000 $\pm$ 0.0000 & / & 0.5754 $\pm$ 0.0000 & 5.6191 $\pm$ 0.0003 & 0.9090 $\pm$ 0.0000 \\
 \specialrule{0em}{1pt}{1pt}
\hline
\specialrule{0em}{1pt}{1pt}
\multirow{4}{*}{\textbf{MGTAB-L}} & 0.0000 & 0.833 $\pm$ 0.000 & 0.4995 $\pm$ 0.0000 & 39593.0 $\pm$ 0.0 & 0.0001 $\pm$ 0.0000  & / & 0.9206 $\pm$ 0.0000 & 2.8082 $\pm$ 0.0000 & 0.8000 $\pm$ 0.0000\\
& 0.0200 &0.716 $\pm$ 0.012 & 0.5045 $\pm$ 0.0000 & 40263.0 $\pm$ 2.8 &  0.0001 $\pm$ 0.0000 & /  &  0.9214 $\pm$ 0.0000 & 2.8068 + 0.0001  &   0.7974 + 0.0001 \\
& 0.0400 & 0.696 $\pm$ 0.047 & 0.5095 $\pm$ 0.0000  & 40267.0 $\pm$ 3.4 & 0.0001 $\pm$ 0.0000 & / & 0.9221 $\pm$ 0.0000 & 2.8060 $\pm$ 0.0004 &  0.7947 $\pm$ 0.0003\\
& 0.0600  & 0.718 $\pm$ 0.007 & 0.5145 $\pm$ 0.0000 & 40267.6 $\pm$ 2.4 & 0.0000 $\pm$ 0.0000 & /& 0.9229 $\pm$ 0.0000 & 2.8051 $\pm$ 0.0005 & 0.7920 $\pm$ 0.0003 \\
\bottomrule
\hline
\end{tabular}}}
\end{sidewaystable}

\begin{figure*}[t]
    \centering
    \subfloat[MAGTAB-L]{
    \centering
    \includegraphics[width=0.33\textwidth]{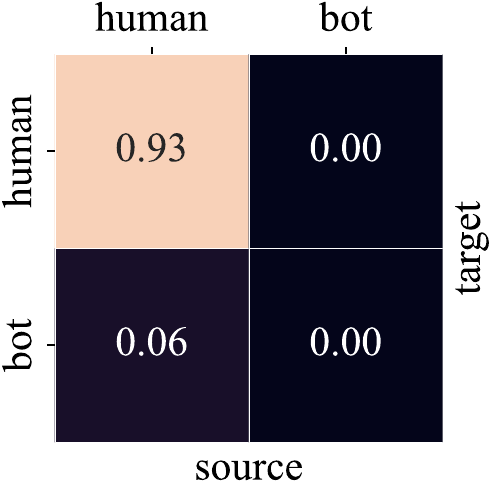}
    }
    \subfloat[MAGTAB-L-ATT]{
    \centering
    \includegraphics[width=0.33\textwidth]{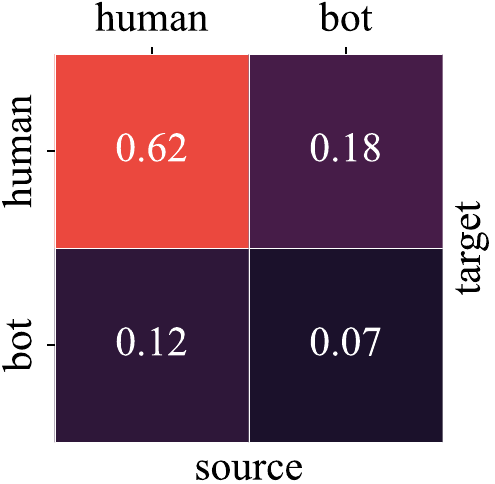}
    }
    \subfloat[Cresci-15]{
    \label{att-cresci-15-fig}
    \centering
    \includegraphics[width=0.33\textwidth]{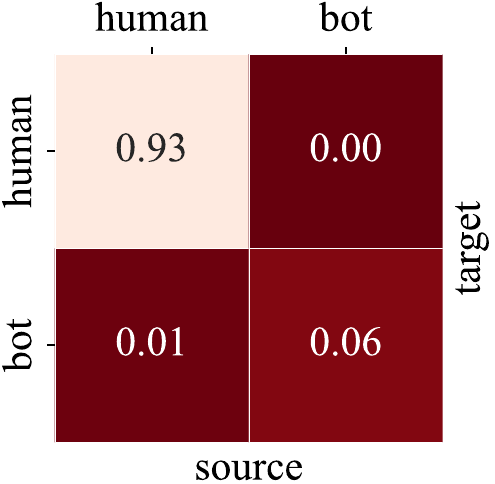}
    }
    \\
    \subfloat[Cresci-15-ATT]{
    \label{att-cresci-15-fig-att}
    \centering
    \includegraphics[width=0.33\textwidth]{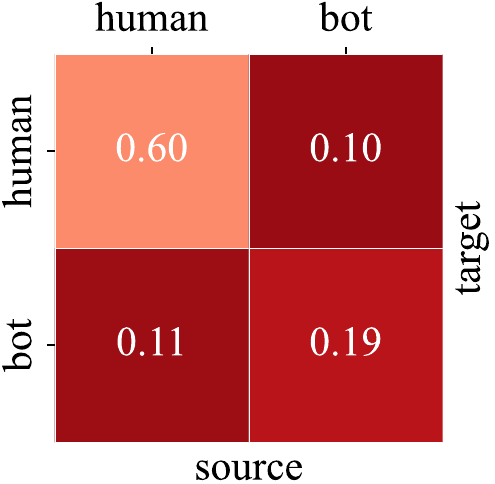}
    }
    \subfloat[MAGTAB]{
    \centering
    \includegraphics[width=0.33\textwidth]{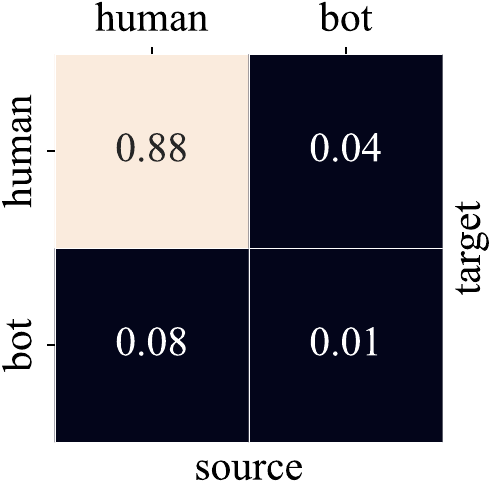}
    }
    \subfloat[MAGTAB-ATT]{
    \centering
    \includegraphics[width=0.33\textwidth]{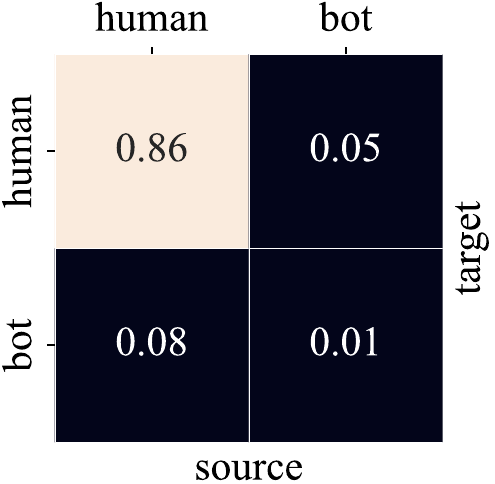}
    }
    \centering
    \caption{The proportion of source node and target node types before and after the \RoBCtrl's attack.}
    \label{fig:visualization}
    
\end{figure*}


\begin{figure}[t]
    \centering
    \subfloat[Cresci-15]{
    \centering
    \includegraphics[width=0.4\textwidth]{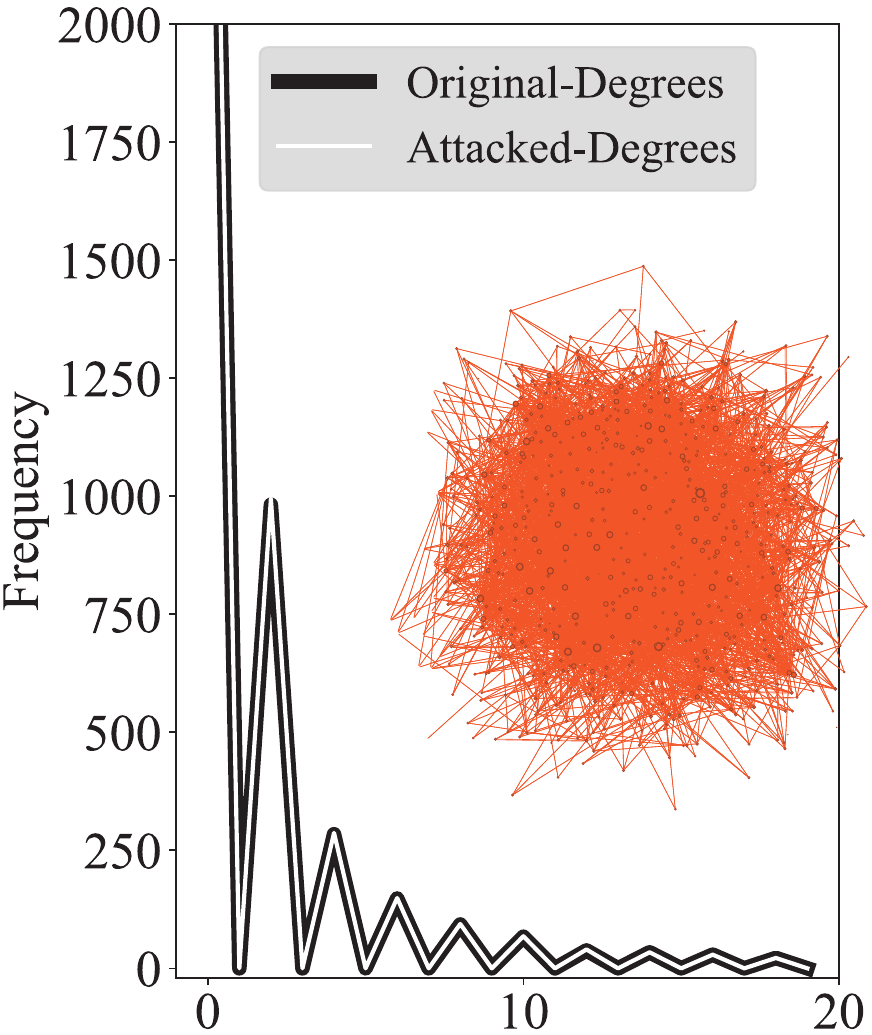}
    }
    \subfloat[MAGTAB]{
    \centering
    \includegraphics[width=0.4\textwidth]{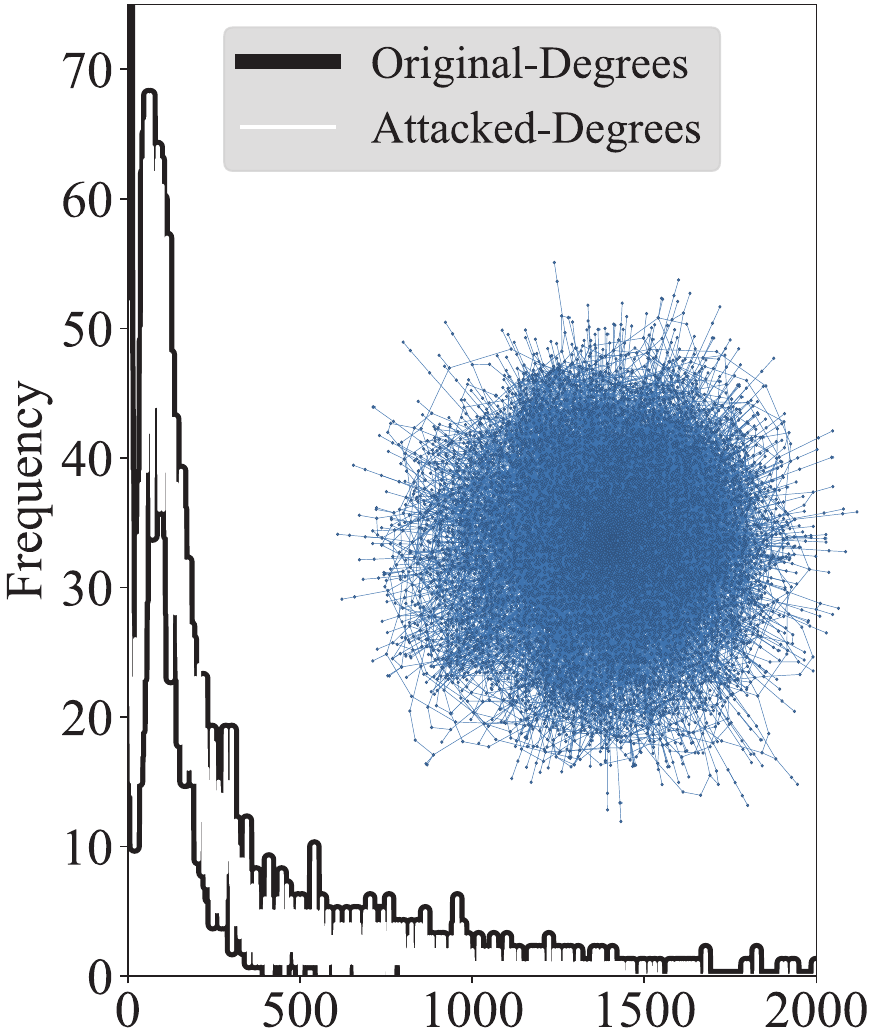}
    }  
    \caption{Distributions of node degrees before and after attack. (Note: The white line always exists inside the black line.)}
    \label{fig:visual_distribution}
\end{figure}
\subsubsection{Performance Comparison on the Large-scale Social Graph}
On social network platforms, adversaries inherently lack knowledge about the underlying architecture of the detection model. 
Consequently, an auxiliary surrogate model $\mathcal{M}$ (in this experiment, a GCN was used) must be employed to facilitate the strategic design of attacks targeting the social graph's complex structure. 
In the subsequent stage, the constructed poisoned graph is used to directly attack the test phase of the deployed model, aiming to degrade its performance.
During the execution of the poisoning attack, the attack budget serves as a governing mechanism, regulating the number of adversarial edges injected into the graph. 
To provide each labeled node with additional connections and maintain a similar level of density between the poisoned and clean graphs, we set the attack budget to be between one and three times the number of labeled nodes, i.e., $\Delta_e = r *|E| \in (|Y|, 3|Y|)$. 
In this context, $r$ signifies the proportion of the injected edges relative to the total edge count.
For the \textbf{TwiBot-20} and \textbf{MGTAB-Large} datasets, we assigned $r=0.04$.
In the \textit{RoBCtrl} attacking scenario, we kept the bots count generated by \DiffBot at a constant 50---a quantity significantly lower than the node count inherent to the social graph. 
This action was executed to avoid inadvertently modifying the graph's intrinsic attributes.
Table~\ref{tab:comparision-table-large} reports the attack performance of \RoBCtrl compared with baseline methods on \textbf{TwiBot-20} and \textbf{MGTAB-Large}. 
The \textbf{Clean} row within the table reports the accuracy rates of various detection models for node classifications on the unpoisoned versions of the social graph datasets. 
In this study, the scale of the graph and the attack type posed significant challenges for attack methods that rely on gradient-based optimization of the adjacency matrix. (Notably, the \textbf{Topology} method could not be applied to the MGTAB-Large dataset due to its substantial memory and GPU requirements, which exceed the capacity of the server used in the experiments). 
Consequently, we only evaluated three baseline methods.
As Table~\ref{tab:comparision-table-large} shows, all attack methods effectively decreased the accuracy of the detection model on the large graphs, which aligns with our expectations.
Notably, \RoBCtrl caused a greater performance drop for each detector on large graphs compared to other strategies, as indicated in the last row of the table. 
We also observed that the performance of the SAGE detection method remained relatively stable. 
Upon analysis, we found that although a few poisoned edges existed in these large-scale graphs, the SAGE method's neighbor-sampling mechanism discarded these proportionally minor poisoned edges during the GNN's message aggregation phase, thereby reducing the attack method's potential impact.

\begin{sidewaystable}[thp]

    \setlength{\belowcaptionskip}{16.cm}
    \aboverulesep=0ex
    \belowrulesep=0ex
    \centering
    \renewcommand\arraystretch{1.3}

\caption{Accuracy and attack time results after attacked by \RoBCtrl and State Abstraction in normal-scale social bot graph datasets.}
\label{tab: sa-statistics-table}
\centering
\scalebox{0.87}{
\setlength{\tabcolsep}{0.05mm}{
\begin{tabular}{c|cc|cc|cc|cc|cc|cc}
\hline
\toprule
\specialrule{0em}{1pt}{1pt}
\multirow{3}{*}{\textbf{Detectors}} & \multicolumn{12}{c}{\textbf{Cresci-15}} \\
\specialrule{0em}{1pt}{1pt}
\cline{2-6}\cline{7-13}
\specialrule{0em}{1pt}{1pt}
& \multicolumn{2}{c}{GCN} & \multicolumn{2}{c}{GAT} & \multicolumn{2}{c}{SAGE} & \multicolumn{2}{c}{AirGNN} & \multicolumn{2}{c}{APPNP} & \multicolumn{2}{c}{SGC} \\ 
\specialrule{0em}{1pt}{1pt}
\hline
\specialrule{0em}{1pt}{1pt}
\textbf{Methods} & Accuracy & Time & Accuracy & Time & Accuracy & Time & Accuracy & Time & Accuracy & Time & Accuracy & Time \\
\specialrule{0em}{1pt}{1pt}
\hline
\specialrule{0em}{1pt}{1pt}
\textbf{Clean} & 92.21+1.02 & / & 91.34+0.64 & / & 94.45+0.32 & / & 94.21+0.55 & / & 94.45+0.32 & / & 91.65+0.08 & / \\
\specialrule{0em}{1pt}{1pt}
\hline
\specialrule{0em}{1pt}{1pt}
\textbf{\RoBCtrl} & 74.02+5.28 & 9.30+0.83 & 78.43+2.63 & 9.23+0.50 & 79.72+3.97 & 9.34+0.25 & 76.76+3.33 & 8.06+1.05 & 79.75+3.36 & 8.46+0.57 & 81.37+7.30 & 8.82+0.67 \\
\textbf{\RoBCtrlSA} & \textbf{72.40}+3.99 & \textbf{8.02}+0.02 & \textbf{77.24}+0.72 & \textbf{8.83}+0.31 & \textbf{78.66}+2.81 & \textbf{9.16}+0.06 & \textbf{74.58}+0.89 & \textbf{7.79}+0.02 & \textbf{78.50}+2.26 & \textbf{8.00}+0.25 & \textbf{79.81}+6.14 & \textbf{8.47}+0.50 \\
\specialrule{0em}{1pt}{1pt}
\hline
\specialrule{0em}{1pt}{1pt}
\textbf{Drop $\downarrow$} & 19.81 & / & 14.1 & / & 15.79 & / & 19.63 & / & 15.95 & / & 11.84 & / \\
\specialrule{0em}{1pt}{1pt}
\hline
\specialrule{0em}{1pt}{1pt}
\textbf{Improve $\uparrow$} & 1.62(2.19\%) & 1.28 (13.76\%) & 1.19 (1.52\%) & 0.4 (4.33\%) & 1.06 (1.33\%) & 0.18 (1.93\%) & 2.18 (2.84\%) & 0.27 (3.35\%) & 1.25 (1.57\%) & 0.46 (5.44\%) & 1.56 (1.92\%) & 0.35 (3.97\%) \\
\specialrule{0em}{1pt}{1pt}
\hline
\specialrule{0em}{1pt}{1pt}
\multirow{3}{*}{\textbf{Detectors}} & \multicolumn{12}{c}{\textbf{MAGTAB}} \\
\specialrule{0em}{1pt}{1pt}
\cline{2-6}\cline{7-13}
\specialrule{0em}{1pt}{1pt}
& \multicolumn{2}{c}{GCN} & \multicolumn{2}{c}{GAT} & \multicolumn{2}{c}{SAGE} & \multicolumn{2}{c}{AirGNN} & \multicolumn{2}{c}{APPNP} & \multicolumn{2}{c}{SGC} \\
\specialrule{0em}{1pt}{1pt}
\hline
\specialrule{0em}{1pt}{1pt}
\textbf{Methods} & Accuracy & Time & Accuracy &Time & Accuracy &Time & Accuracy & Time & Accuracy & Time & Accuracy & Time \\
\specialrule{0em}{1pt}{1pt}
\hline
\specialrule{0em}{1pt}{1pt}
\textbf{Clean} & 77.89+6.16 & / & 84.43+0.12 & / & 86.10+0.65 & / & 82.42+5.96 & / & 84.27+0.47 & / & 75.17+0.53 & / \\
\specialrule{0em}{1pt}{1pt}
\hline
\specialrule{0em}{1pt}{1pt}
\textbf{\RoBCtrl} & 66.74+6.25 & 7.70+0.62 & 74.22+1.09 & 7.71+0.48 & 75.48+0.47 & 8.59+0.05 & 71.48+10.09 & 7.27+1.01 & 52.21+14.54 & 6.04+0.87 & 72.00+2.91 & 7.22+0.29 \\
\textbf{\RoBCtrlSA} & \textbf{65.45}+4.90 & \textbf{7.54}+0.49 & \textbf{72.31}+0.31 & \textbf{7.33}+0.07 & \textbf{74.15}+0.65 & \textbf{8.45}+0.06 & \textbf{65.10}+0.86 & \textbf{6.51}+0.05 & \textbf{49.95}+8.66 & \textbf{5.93}+0.75 & \textbf{70.17}+0.26 & \textbf{7.02}+0.01 \\
\specialrule{0em}{1pt}{1pt}
\hline
\specialrule{0em}{1pt}{1pt}
\textbf{Drop $\downarrow$} & 12.44 & / & 12.12 & / & 11.95 & / & 17.32 & / & 34.32 & / & 5.00 & / \\
\specialrule{0em}{1pt}{1pt}
\hline
\specialrule{0em}{1pt}{1pt}
\textbf{Improve $\uparrow$} & 1.29 (1.93\%) & 0.16 (2.08\%) & 1.91 (2.57\%) & 0.38 (4.93\%) & 1.33 (1.76\%) & 0.14 (1.63\%) & 6.38 (8.93\%) & 0.76 (10.45\%) & 2.26 (4.33\%) & 0.11 (1.82\%) & 1.83 (2.61\%) & 0.2 (2.77\%) \\

\bottomrule
\hline
\end{tabular}}
}

\end{sidewaystable}

\subsubsection{Performance Comparison on Normal-Scale Social Graph}

Two datasets were used to test the attack efficiency of \RoBCtrl on normal-scale social graphs. 
Additionally, we selected two target attack baseline methods, FGA and IGA, to compare the performance of the target attack methods with that of our proposed method.
We set the attack ratio as $r=0.15$ for Cresci-15 and $r=0.0035$ in MGTAB. Table \ref{tab:normal-scale-comparision-table} reports the attacking results. Even on normal-scale graphs, \RoBCtrl retained a discernible edge in performance over the other baseline attack methods. Moreover, our method substantially degraded detector performance on Cresci-15; however, the impact on MGTAB was more mitigated. This variation can be attributed to the fewer interactions among the accounts within the Cresci-15 dataset. As a result, the insertion of just a few edges can substantially affect the message aggregation mechanism in GNN-based bot detectors. At the same time, it is evident that the detection performance of the \textbf{SAGE} method, even with sampling, was significantly impacted in this experiment. In contrast, the MGTAB dataset features an extremely dense social graph due to the high volume of user interactions, as detailed in Table~\ref{tab:statistics}. Therefore, the added poison edges represent only a small fraction of the total number of edges, exerting minimal influence on the detector’s operational performance.

\subsubsection{Attack Combination Strategy (Ablation study)}
To handle the scenario in which an attacker deploys a social bots army, distinct generations of social bots are considered because of the variability in the influence and costs associated with each bot account. To model diverse attack strategies, we established two variants---\RoBCtrl-1 and \RoBCtrl-2---configured in tandem with RoBCtrl. \RoBCtrl-1 refers to the deployment of solely automated bots, whereas \RoBCtrl-2 is a combined deployment of both automated and cyborg bots. As the results in Tables~\ref{tab:comparision-table-large} and \ref{tab:normal-scale-comparision-table} reveal, control over low-cost automated bots alone can substantially impact the detector's accuracy with regard to the target accounts. Nevertheless, the integration of cyborg and evolved bots, as emulated by the diffusion model, can further compromised the detector's accuracy. This implies the higher efficacy of our attack strategy, which consists of simultaneous control over diverse bot generations. 

\begin{figure}[t]
    \centering
        \centering
        \includegraphics[width=0.0215\textwidth]{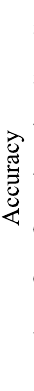}
        \captionsetup{labelformat=empty}
        \hspace{-0.3cm}
        \subfloat[GCN as defense method on Cresci-15]{        
        \centering
        \includegraphics[width=0.45\textwidth]{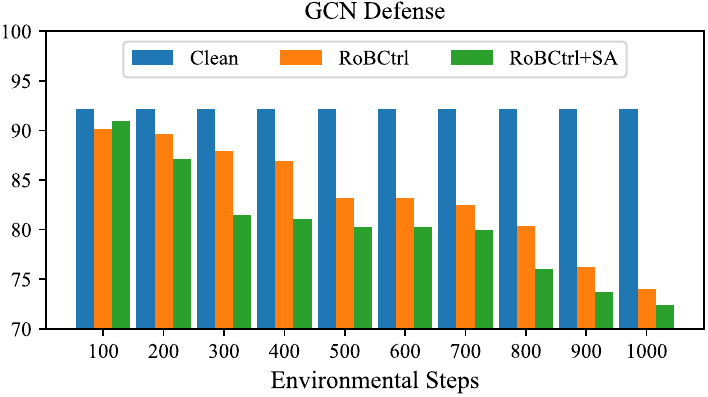}
        \label{subfig:cresci_sa1}
        }
        \centering
        \includegraphics[width=0.0215\textwidth]{assets/magtabsa_title.pdf}
        \captionsetup{labelformat=empty}
        \hspace{-0.3cm}
        \subfloat[GCN as defense method on MAGTAB]{
        \centering
        \includegraphics[width=0.45\textwidth]{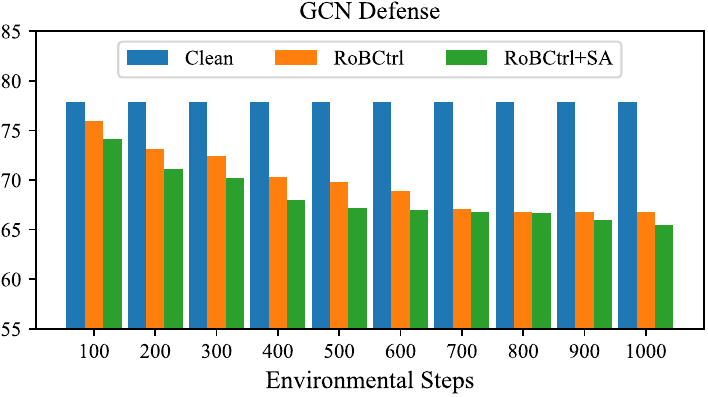}
        \label{subfig:magtab_sa1}
        }

        \centering
        \includegraphics[width=0.0215\textwidth]{assets/magtabsa_title.pdf}
        \captionsetup{labelformat=empty}
        \hspace{-0.3cm}
        \centering
        \subfloat[GAT as defense method on Cresci-15]{
        \centering
        \includegraphics[width=0.45\textwidth]{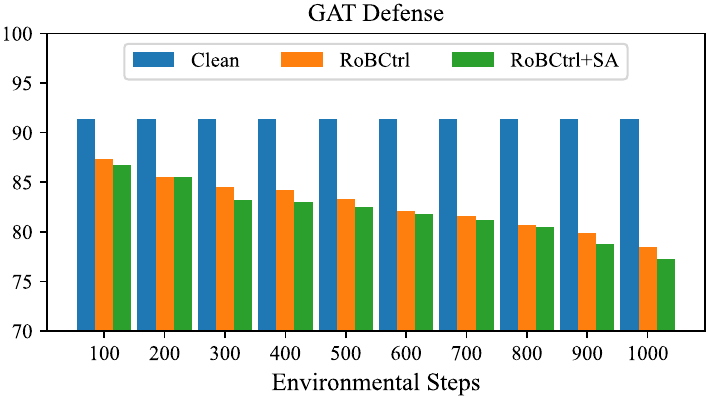}
        \label{subfig:cresci_sa2}
        }
        \centering
        \includegraphics[width=0.0215\textwidth]{assets/magtabsa_title.pdf}
        \captionsetup{labelformat=empty}
        \hspace{-0.3cm}
        \subfloat[APPNP as defense method on MAGTAB]{
        \centering
        \includegraphics[width=0.45\textwidth]{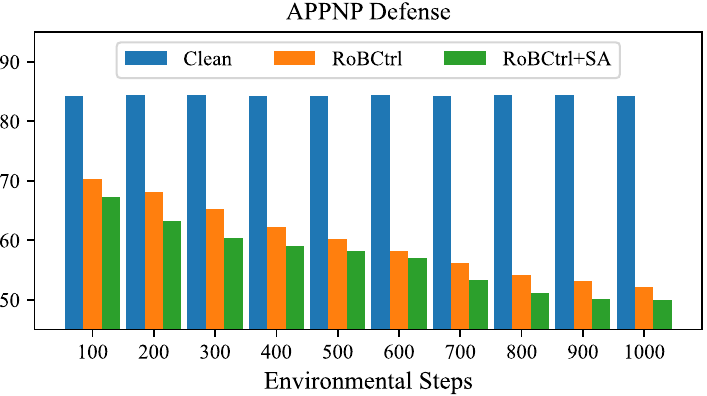}
        \label{subfig:magtab_sa2}
        }

        \centering
        \includegraphics[width=0.0215\textwidth]{assets/magtabsa_title.pdf}
        \captionsetup{labelformat=empty}
        \hspace{-0.3cm}
        \centering
        \subfloat[AirGNN as defense method on Cresci-15]{
        \centering
        \includegraphics[width=0.45\textwidth]{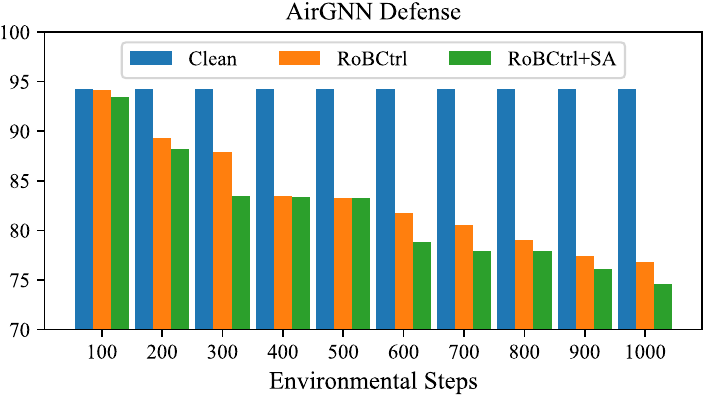}
        \label{subfig:cresci_sa3}
        }
        \centering
        \includegraphics[width=0.0215\textwidth]{assets/magtabsa_title.pdf}
        \captionsetup{labelformat=empty}
        \hspace{-0.3cm}
        \subfloat[AirGNN as defense method on MAGTAB]{
        \centering
        \includegraphics[width=0.45\textwidth]{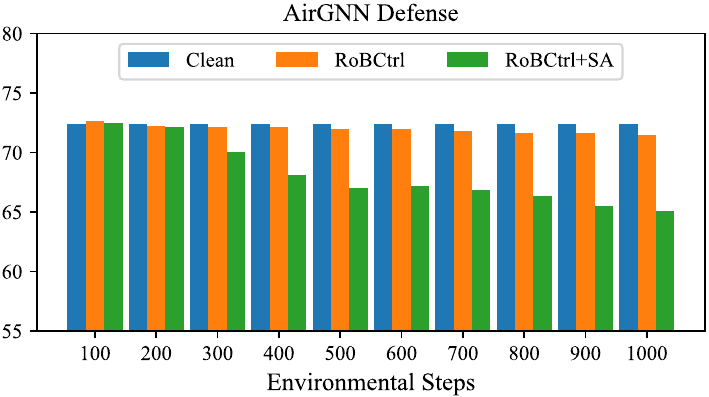}
        \label{subfig:magtab_sa3}
        }
    
    \caption{Attack performance of the \RoBCtrl framework, with and without state abstraction at different training steps, on the Cresci-15 (\textit{left}) and MAGTAB (\textit{right}) datasets.}
    \label{fig:training process}
\end{figure}
\subsection{RQ2: Statistics of the Poisoned Graphs }
\label{exp:Q2}
To answer \textbf{RQ2}, we examine several critical statistics of the compromised social networks to gain insights into the nature of the attack behavior. Ideally, a poisoned graph should largely retain the statistical characteristics of its clean counterpart. In this study, we employed key graph statistics used in prior work to evaluate the poisoned graphs across four datasets. Specifically, we report measures such as the accuracy, average degree, the largest connected component, clustering coefficient, Gini coefficient, power Law exponent, and distribution entropy to thoroughly analyze characteristics of the poisoned graphs, such as distribution and density. Detailed definitions and descriptions of these measures can be found in \cite{bojchevski2018netgan}. The resultant outcomes are presented in Table \ref{tab:statistics-table}. Preliminary analysis suggests that the accuracy of the detector diminished proportionally with an increase in attack disturbances. Interestingly, the decrease in accuracy plateaus at a certain point, suggesting that there is a limit to the number of edges that can impact the detector's accuracy. Upon further observation, we note that the poisoned graph maintained a distribution remarkably similar to that of the clean graph. Indications of this similarity are evident in corresponding measures such as the power law exponents and distribution entropy, suggesting that both the poisoned and clean versions share similar distribution patterns.

\subsection{RQ3: Attack Strategy Evaluation}
To address \textbf{RQ3}, we conducted an analysis of the test set's connected edges in the social graph before and after the \RoBCtrl attack. As shown in Figure \ref{fig:visualization}, we examined the percentage of edges between different node types. In the original social graph, real user accounts exhibited numerous interactions among themselves. In Figures \ref{att-cresci-15-fig} and \ref{att-cresci-15-fig-att}, bot accounts have more connections with other bot accounts. However, after the attack, there was a noticeable increase in the connections between real users and bot accounts, which reduced detector performance. This finding further confirms the effectiveness of the attack strategy, wherein bot accounts interact more with real users to evade platform monitoring.
We also analyzed the distribution of node degree in the dataset before and after the attack, as depicted in Figure \ref{fig:visual_distribution}. We observed only slight variations in the degree distribution of the graph, indicating the reliability of our attack method in preserving the graph's original properties and the invisibility of attack by social bot accounts in terms of node degree distribution. On the basis of these experimental results, we recommend shifting the focus of countermeasures against social account attacks from individual account detection to group-level detection, and from individual account features to group account feature detection. This approach could be instrumental in detecting large-scale coordinated actions of social accounts. Additionally, our method demonstrates the potential of adversarial learning or techniques to enhance the robustness of detectors.

\subsection{RQ4: State Abstraction}
To address \textbf{RQ4}, we have compared the attack performance of two variants, \textbf{\RoBCtrlSA} (with the state abstraction mechanism) and \textbf{\RoBCtrl} (without the state abstraction mechanism), using the \textbf{Cresci-15} and \textbf{MAGTAB} datasets, as described in Section \ref{section: stateabstraction}. 
In this evaluation, we fixed the training steps to 1000 and employed the trained models to execute attacks against all six black-box detectors.

Table \ref{tab: sa-statistics-table} summarizes the difference in accuracy (\%) and time spent (ms) for each attack, representing the attack's effectiveness and efficiency, respectively. 
In terms of effectiveness, the \textbf{\RoBCtrlSA} framework reduced the accuracy of each detector, with an average drop of 16.19\% in \textbf{Cresci-15} and 15.52\% in \textbf{MAGTAB}. 
In terms of efficiency, as shown in the \textbf{Improve} rows, the \textbf{\RoBCtrlSA} variant outperformed \textbf{\RoBCtrl} in attack time, with a reduction of 5.53\% in \textbf{Cresci-15} and 3.93\% in \textbf{MAGTAB} on average.

Moreover, for each dataset, we selected three defense methods that exhibited the largest drop in accuracy drop after the attack and reported their accuracies before and after the attacks at different steps (Figure \ref{fig:training process}). 
Regardless of the number of environment steps used for training, the \textbf{\RoBCtrlSA} variant consistently executed more effective attacks against each graph-based detector, further demonstrating the higher effectiveness and efficiency advantages provided by the state abstraction mechanism.
\section{CONCLUSION AND DISCUSSION}
\label{sec:Conclusion}
In this paper, we investigate the vulnerability of GNN-based social bots detectors under structured adversarial attacks. We primarily use a diffusion model to simulate the generation process of advanced social bots and employ a MARL framework to mimic strategies for evading detection. 
We hope our findings will inform the development of effective defenses, enabling better protection against adversarial threats.
Extensive experiments on social bot detection datasets demonstrate that our method can effectively disrupt social network ecosystems and degrade the performance of GNN-based detectors. 
Furthermore, additional experiments on two datasets show that state abstraction can significantly reduce the time required to execute successful attacks. 
Our results highlight the potential threats posed by adversarial attacks and aim to pave the way for developing stronger defenses. 
By identifying weaknesses in current detection systems, we strive to contribute to more robust solutions for safeguarding against such attacks.

\textbf{Limitation and Future work.} 
Despite the contributions of this work, several limitations should be noted. 
Our primary focus has been developing and demonstrating a practical adversarial attack framework for GNN-based social bot detectors, emphasizing the vulnerabilities of existing systems. 
However, this study does not explore potential defense mechanisms or protection strategies against such attacks. 
Although understanding the limitations of current detectors is a critical first step toward enhancing their robustness, the lack of discussion of mitigation strategies, such as adversarial training, robust model architectures, or detection of adversarial patterns, leaves an essential aspect of the problem unaddressed. 
Future work aims to close this gap by investigating comprehensive defense frameworks that counteract the types of attack presented here and preemptively safeguard against emerging threats.

\textbf{Broader Impact.}
This work provides valuable insights into the vulnerabilities of GNN-based social bot detectors, which are widely used to identify malicious social media activities. 
By revealing the susceptibility of these systems to structured adversarial attacks, this research underscores the need for more robust detection models, contributing to the broader goal of enhancing the integrity of online social ecosystems.
However, the potential misuse of the proposed adversarial framework for unethical purposes poses a significant concern. 
While we intend to advance the development of stronger defenses against adversarial threats, there is a risk that malicious actors could exploit these insights to refine their attack strategies. 
To mitigate such risks, this research emphasizes the ethical responsibility of its use and advocates for developing proactive defense mechanisms as a critical next step. 
Overall, this work aims to strengthen the resilience of social platforms, ultimately promoting safer and more trustworthy digital environments.

\textbf{Ethical Statement.}
While this work proposes an adversarial attack framework for graph-based social bot detectors, we intend to focus on detecting and enhancing the robustness of existing detectors. Therefore, we do not endorse the utilization of this work for unethical purposes in any manner.


\bibliographystyle{ACM-Reference-Format}
\bibliography{sample-base}

\appendix

\end{document}